\documentclass[preprint,12pt]{elsarticle}
\usepackage[hidelinks]{hyperref}
\usepackage{amssymb}
\usepackage[margin=1.1in]{geometry}
\usepackage{amsmath}
\usepackage{tabularx}
\usepackage[dvipsnames]{xcolor}
\usepackage{graphicx}
\usepackage{subcaption}
\usepackage{setspace}
\usepackage[utf8]{inputenc}
\usepackage{pgfplots}
\usetikzlibrary{quotes,arrows.meta}
\usepgfplotslibrary{groupplots}
\usetikzlibrary{shapes.geometric}
\usepgfplotslibrary{groupplots,dateplot}
\usetikzlibrary{patterns,shapes.arrows}
\pgfplotsset{
  compat=newest
}

\pgfdeclareplotmark{mystar}{
    \node[star,star point ratio=2.25,minimum size=10pt,
          inner sep=0pt,draw=black,solid,fill=red] {};
}

\usepackage{algorithm}
\usepackage{algorithmic}
\usepackage[hidelinks]{hyperref}
\usepackage{siunitx}
\usepackage{amsfonts}
\usepackage{bm}
\usepackage{caption}
\newcommand{\Bdobs}{\mathbf{d}_\text{obs}}

\journal{Advances in Water Resources}

\begin{document}

\begin{frontmatter}

\title{Likelihood-Free Inference and Hierarchical Data Assimilation for Geological Carbon Storage}
\author[a]{Wenchao Teng}
\author[a]{Louis J.~Durlofsky}

\affiliation[a]{{Department of Energy Science and Engineering, Stanford University, Stanford, CA 94305, USA}}
\begin{abstract}
Data assimilation will be essential for the management and expansion of geological carbon storage operations. In traditional data assimilation approaches a fixed set of geological hyperparameters, such as mean and standard deviation of log-permeability, is often assumed. Such hyperparameters, however, may be highly uncertain in practical CO$_2$ storage applications where measurements are scarce. In this study, we develop a hierarchical data assimilation framework for carbon storage that treats hyperparameters as uncertain variables characterized by hyperprior distributions. To deal with the computationally intractable likelihood function in hyperparameter estimation, we apply a likelihood-free (or simulation-based) inference algorithm, specifically sequential Monte Carlo-based approximate Bayesian computation (SMC-ABC), to draw posterior samples of hyperparameters given dynamic monitoring well data. In the second step we use an ensemble smoother with multiple data assimilation (ESMDA) procedure to provide posterior realizations of grid-block permeability. To reduce computational costs, a 3D recurrent R-U-Net deep learning-based surrogate model is applied for forward function evaluations. The accuracy of the surrogate model is established through comparisons to high-fidelity simulation results. A rejection sampling (RS) procedure for data assimilation is applied to provide reference posterior results. Detailed data assimilation results from SMC-ABC-ESMDA are compared to those from the reference RS method. These include marginal posterior distributions of hyperparameters, pairwise posterior samples, and history matching results for pressure and saturation at the monitoring location. Close agreement is achieved with ‘converged’ RS results, for two synthetic true models, in all quantities considered. Importantly, the SMC-ABC-ESMDA procedure provides speedup of 1--2 orders of magnitude relative to RS for the two cases. A modified standalone ESMDA procedure, able to treat uncertain hyperparameters, is introduced for comparison purposes. For the same number of function evaluations, the hierarchical data assimilation approach is shown to provide superior results for posterior hyperparameter distributions and monitoring well pressure predictions.

\end{abstract}


\end{frontmatter}

\section{Introduction}
Geological carbon storage offers a potential means to significantly reduce $\mathrm{CO}_{2}$ emissions to the atmosphere. Such operations, however, typically involve substantial uncertainty due to the limited information available regarding storage aquifer properties. This uncertainty can be reduced by applying data assimilation (or history matching) procedures, which act to provide geological realizations that give flow simulation results in approximate agreement with observed data. A common issue in traditional data assimilation frameworks, however, is the challenge of determining appropriate values for the relevant hyperparameters (also referred to as scenario parameters or metaparameters). Examples of these hyperparameters are the mean and standard deviation of porosity and permeability, permeability anisotropy ratio, correlation lengths, etc. Because inappropriate assumptions regarding hyperparameter values can lead to implausible posterior realizations and biased predictions~\cite{oliver2018calibration}, it is important to include hyperparameter estimation in the data assimilation procedure.

In this paper, we present a hierarchical data assimilation framework to estimate hyperparameters and grid-block permeability values given dynamic monitoring well data. In contrast to many traditional data assimilation approaches, we treat the hyperparameters as uncertain. Our specific procedure involves a sequential Monte Carlo-based approximate Bayesian computation (SMC-ABC) method for hyperparameter estimation, followed by an ensemble smoother with multiple data assimilation (ESMDA) to provide posterior realizations of grid-block permeability. These treatments require a substantial number of function evaluations. We thus apply a deep learning-based flow simulation surrogate model \cite{tangmeng2022Deep, han2024surrogate} in place of a high-fidelity numerical simulator. The use of this surrogate model also enables us to compare data assimilation results from our method with those from the much more computationally intensive rejection sampling procedure.

Data assimilation has been extensively studied and applied in subsurface flow settings. Ensemble Kalman-based approaches, such as ensemble Kalman filter (EnKF)~\cite{evensen2003ensemble} and ensemble smoother with multiple data assimilation (ESMDA)~\cite{emerick2013ensemble}, are widely used for oil reservoir simulation and for geological carbon storage (GCS) modeling. In the context of GCS, ensemble-based methods have been applied, for example, for the detection of potential leakage pathways~\cite{gonzalez2015detection} and for uncertainty reduction in risk assessment studies~\cite{chen2020reducing}. Different types of measurements have been considered with ensemble-based methods. These include time-lapse seismic surveys~\cite{liu2020petrophysical}, microseismic events~\cite{jahandideh2021inference}, and InSAR surface displacement data~\cite{tang2022deep}. The large volume of observed data has also stimulated the application of more advanced techniques for dimension reduction and efficient parameterization~\cite{liu2020petrophysical,tang2022deep}. The above-mentioned studies did not, however, consider hyperparameter uncertainty.

Deep learning-based surrogate models have been actively developed for GCS applications. Our discussion on this topic will be brief, however, as the development of a surrogate model is not the focus of this work. A range of network architectures has been used for GCS modeling. These include the Recurrent Residual U-Net~\cite{tangmeng2021deep}, Wide ResNet~\cite{tang2021deep}, Fourier Neural Operators~\cite{wen2022u, seabra2024ai}, and Transformer U-Net~\cite{seabra2024ai}. The use of these surrogate models has enabled formal, though computationally demanding, data assimilation approaches to be applied for carbon storage problems. Tang et al.~\cite{tangmeng2022Deep}, for example, used their 3D recurrent R-U-Net surrogate model with a rejection sampling method to history match coupled flow and geomechanics problems. Han et al.~\cite{han2024surrogate} extended the recurrent R-U-Net model and incorporated it into a Markov chain Monte Carlo (MCMC) history matching procedure in which hyperparameter uncertainty was also considered.

Compared to traditional data assimilation (with assumed/fixed hyperparameters), less effort has been directed toward the hierarchical Bayesian inference problem~\cite{gelman1995bayesian} involving the joint estimation of posterior hyperparameters and geological realizations. One of the first approaches to address this issue was the quasi-linear geostatistical theory by Kitanidis~\cite{kitanidis1995quasi}. Based on this work, Zhao and Luo~\cite{zhao2021bayesian} proposed an iterative method to correct the prior hyperparameters. These procedures rely on linearization and thus have limitations for strongly nonlinear problems. MCMC has also been used for hierarchical Bayesian problems in subsurface settings~\cite{hansen2012inverse, reuschen2020bayesian, xiao2021bayesian, han2024surrogate}. Most MCMC approaches are expensive, however, and they do not naturally parallelize. Some of these methods involve a hierarchically noncentered parameterization in which hyperparameters and latent variables are a priori independent~\cite{papaspiliopoulos2007general}. Such a parameterization is also applicable for ensemble Kalman-based data assimilation when the augmented parameter space is assumed to be Gaussian~\cite{chada2018parameterizations, oliver2022hybrid}. The above-mentioned studies using a noncenterd parameterization rely on parameter covariance matrix factorization and latent Gaussian models. These factors introduce additional challenges to data assimilation in 3D subsurface problems.

The rigorous calculation of the likelihood of observed data given the hyperparameters is computationally intractable. This is because the likelihood function corresponds to an integral over all possible geomodel realizations. Various treatments have been introduced to address this issue. Friedli et al.~\cite{friedli2023inference} used a correlated pseudo-marginal Metropolis-Hastings algorithm to estimate the marginal posterior distributions of geostatistical hyperparameters given geophysical data. Wang et al.~\cite{wang2022hierarchical} presented a machine learning-based inversion method to estimate the posterior distributions of hydrological hyperparameters. The problem of an intractable likelihood function also arises in other fields, such as population genetics~\cite{beaumont2002approximate}, ecology~\cite{beaumont2010approximate}, and cosmology~\cite{akeret2015approximate}, and this has motivated the development of various likelihood-free inference algorithms. These can be categorized into two broad classes, as we now discuss.

The first class of methods uses the forward simulator directly during the inference process. A widely used method is approximate Bayesian computation (ABC)~\cite{sisson2018handbook}. Efficient approaches have been developed by combining ABC with MCMC~\cite{marjoram2003markov} and with sequential Monte Carlo (SMC) methods~\cite{sisson2007sequential, toni2009Approximate, beaumont2009adaptive}. The second class of likelihood-free methods uses the simulator to train neural conditional density estimators to provide the likelihood~\cite{tran2017hierarchical} or posterior~\cite{papamakarios2019sequential} of hyperparameters. There have only been a few studies and applications of likelihood-free inference methods in subsurface settings~\cite{cui2018emulator, vrugt2013toward}. To our knowledge, these methods have not yet been considered within the context of GCS.

Our goal in this paper is to develop and apply a hierarchical data assimilation framework for geological carbon storage problems. Specifically, we implement a sequential Monte Carlo-based approximate Bayesian computation (SMC-ABC) method to calibrate the hyperparameters, followed by ESMDA to provide posterior realizations of grid-block permeability. Our hierarchical framework is quite different from existing approaches used for data assimilation in GCS. Specifically, the computations in the SMC-ABC method are inherently parallel. This represents a significant advantage over MCMC methods. In addition, the SMC-ABC approach is fundamentally global. It thus avoids getting ‘stuck’ in local modes of the parameter space, and it does not require a burn-in period. Finally, the proposed procedure does not require a noncentered parameterization. We evaluate the performance of the hierarchical procedure for two synthetic ‘true’ models. The rejection sampling (RS) method is used as a reference for the posterior hyperparameter distributions. The substantial efficiency gains provided by the hierarchical procedure relative to the rigorous RS procedure are quantified. We also evaluate the performance of the hierarchical procedure relative to  a standalone ESMDA approach that is modified to handle uncertain hyperparameters.

This paper proceeds as follows. In Section~\ref{sec:surrogate}, we briefly describe the problem setup and the recurrent R-U-Net surrogate model used in this work. Comparisons of surrogate model predictions to high-fidelity simulation results will also be presented. Next, in Section~\ref{sec:hierarchical}, we develop the hierarchical data assimilation procedure involving SMC-ABC followed by ESMDA. An overview of the RS algorithm is also provided. Detailed data assimilation results for two synthetic true models, including quantitative comparisons to RS, are provided in Section~\ref{sec:results}. A modified ESMDA technique is introduced in Section~\ref{sec:Traditional Data Assimilation}, and comparisons involving this method, RS, and SMC-ABC-ESMDA are presented. We conclude with a summary and suggestions for future work in Section~\ref{sec:conclusion}.
\section{Simulation Setup and Surrogate Model} \label{sec:surrogate}

In this section, we first describe the geological model and simulation specifications. The surrogate model and the training procedure are then discussed. Comparisons of surrogate model predictions to reference simulation results are provided.

\subsection{Geomodel and simulation parameters}
\label{sec:model setup}

\begin{figure}[!ht]
\centering
{\includegraphics{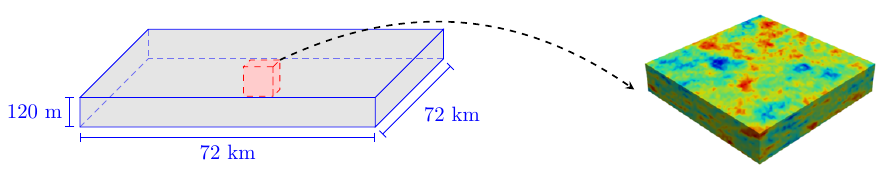}}
\caption{Full simulation domain (left), and central storage aquifer with heterogeneous permeability field (right).}
\label{fig:domain}
\end{figure}

The overall domain considered in the flow simulations, shown in Fig.~\ref{fig:domain}, includes a central storage aquifer and an extensive surrounding region that provides pressure support. The full system is of size 72~km $\times$ 72~km $\times$ 120~m, and the storage aquifer region is 7.2~km $\times$ 7.2~km $\times$ 120~m. The overall system is represented on a grid containing 70 $\times$ 70 $\times$ 12 cells. The storage aquifer grid is 60 $\times$ 60 $\times$ 12. The grid is uniform in the storage aquifer region and nonuniform, with increasing cell dimensions as we move away from the storage aquifer, in the surrounding region. The aquifer is characterized by a heterogeneous permeability field. Porosity is considered to be constant but uncertain. The surrounding region is assigned a constant permeability of 30~md and a constant porosity of 0.2.

Permeability realizations of the storage aquifer are constructed using sequential Gaussian simulation in the geostatistical software library GSLIB~\cite{deutsch1998geostatistical}. The spatial correlation of the permeability field is described using an exponential variogram model. The correlation length (or variogram range) is considered to be an uncertain hyperparameter. The other uncertain hyperparameters considered in this work are the mean and standard deviation of log-permeability, permeability anisotropy ratio, and storage aquifer porosity. The priors of these hyperparameters, referred to as hyperpriors, are all assumed to follow the uniform distributions given in Table~\ref{table_hyperparameters}. Other distributions could also be used in our formulation.

\begin{table}[!ht]
\small
\setstretch{1.5}
\centering
\caption{\normalsize Hyperparameters and hyperpriors}
\label{table_hyperparameters}
\begin{tabular}{p{8cm} p{6.5cm}}
\hline
\noalign{\hrule height 0.8pt}
Parameters & Prior distribution \\ 
\hline
Mean of log-permeability, $\mu_{\log k}$  & $\mu_{\log k} \sim$ $U$(2.5, 4.5) ($\log_e k$, $k$ in md) \\ 
Standard deviation of log-permeability, $\sigma_{\log k}$ & $\sigma_{\log k} \sim$ $U$(0.5, 2) \\ 
Permeability anisotropy ratio, $a_r$  & $\log_{10}a_r$ $\sim$ $U$(-2, 0) \\ 
Horizontal correlation length, $l_h$ & $l_h \sim$ $U$(5, 20) cells\\
Constant porosity, $\phi$ & $\phi$ $\sim$ $U$(0.13, 0.23)\\
\hline
\noalign{\hrule height 0.8pt}
\end{tabular}
\end{table}

\begin{figure}[!ht]
\centering
{\input{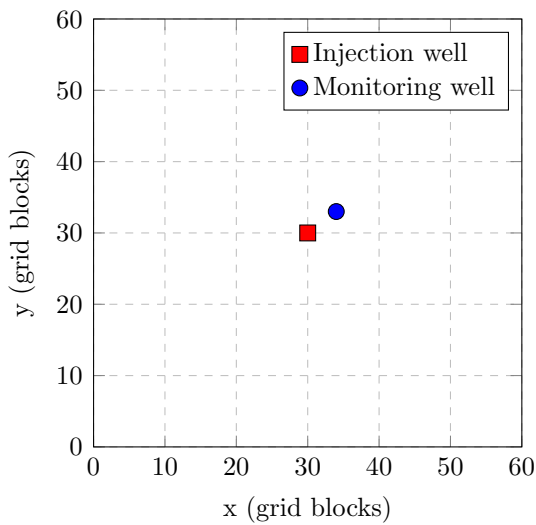}}
\caption{Locations of the injection well and monitoring well in the storage aquifer.}
\label{fig:well_location}
\end{figure}

\begin{figure}[!ht]
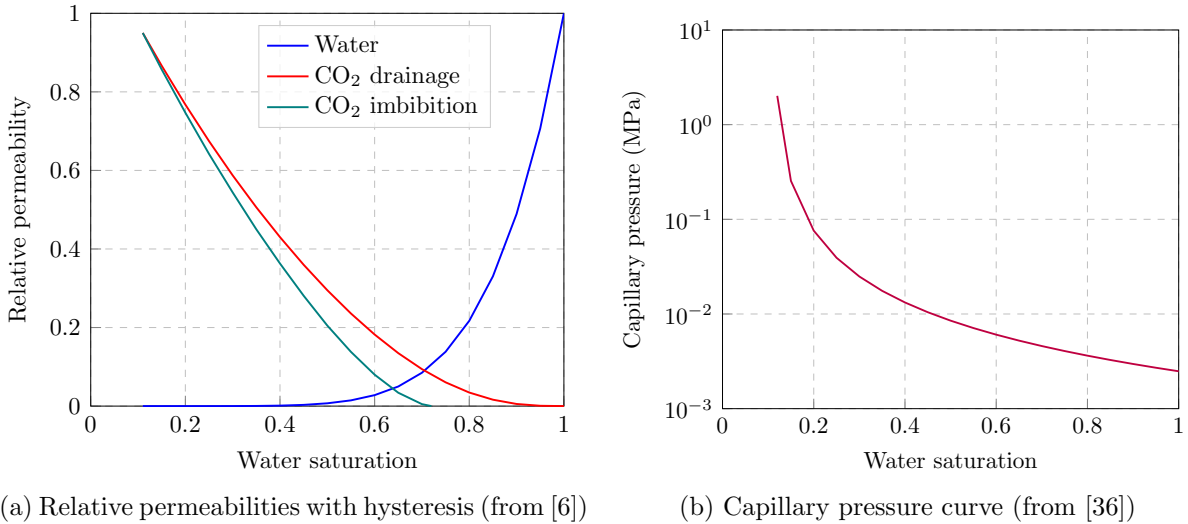

\centering
\begin{subfigure}[t]{0.49\textwidth}
  \centering
  \scalebox{0.87}{\input{Figures/surrogate_model/rel_perm}}
  \caption{Relative permeabilities with hysteresis (from~\citep{cameron2012optimization})}
  \label{fig:rel_perm}
\end{subfigure}
\hfill
\begin{subfigure}[t]{0.49\textwidth}
  \centering
  \scalebox{0.87}{\input{Figures/surrogate_model/capillary}}
  \caption{Capillary pressure curve (from~\citep{sun2019data})}
  \label{fig:capillary}
\end{subfigure}
\caption{\( \text{CO}_{2} \)--brine two-phase flow functions. Curve in (b) is for porosity of 0.2 and permeability of 30~md.}
\label{fig:rel_perm_capillary}
\end{figure}

With the geomodel realizations as input, we perform numerical flow simulation to obtain pressure and saturation fields at a set of time steps. In the simulations, a single fully-penetrating vertical injector is placed in the middle of the storage aquifer. A monitoring well, as shown in Fig.~\ref{fig:well_location}, is located near the injector. We specify a constant $\mathrm{CO}_{2}$ injection rate of 1~Mt/year for 30~years. The aquifer is at an initial pressure of 15.5~MPa at a depth of 1524~m, which corresponds to the top layer of the model. The temperature is constant at 55~$\si{\celsius}$. Relative permeability hysteresis, as shown in Fig.~\ref{fig:rel_perm_capillary}(a), is included in the model. The Leverett J-function is used to calculate heterogeneous capillary pressure based on permeability and porosity. The simulation model does not include any geochemical reactions. All flow simulations are performed using the ECLIPSE simulator with the CO2STORE option~\cite{schlumberger2014eclipse}. Each run requires about 3~minutes (on average) using an AMD EPYC 7502 32-core processor.

\subsection{Surrogate model training and evaluation}

The SMC-ABC-ESMDA methodology developed in this work requires ${\mathcal O}(10^5)$ function evaluations for data assimilation in the examples considered. This method will be compared to a reference rejection sampling procedure that requires ${\mathcal O}(10^6-10^7)$ function evaluations. Given these large simulation demands, it is clearly beneficial to replace the numerical flow simulator with a fast surrogate model for the great majority of function evaluations. A number of accurate and effective such models have been developed (as discussed in the Introduction). Here we will apply an existing framework that is well-suited for our application.

In this work we use the 3D recurrent R-U-Net model presented in~\cite{tangmeng2021deep, tangmeng2022Deep, han2024surrogate}. This model involves convolutional and recurrent (convLSTM) networks, and acts to map the input geomodel to pressure and saturation fields at a specified set of time steps. Separate networks are trained for pressure and saturation variables. The accuracy of recurrent R-U-Net surrogate model predictions has been assessed for $\mathrm{CO}_{2}$ storage problems in~\cite{tangmeng2022Deep,han2024surrogate}, where it was shown to provide results in close agreement (within a few percent) with reference numerical simulation results.

The surrogate model is trained using simulation results for a set of realizations sampled from the prior. This entails drawing 2000 sets of hyperparameters (from the distributions given in Table~\ref{table_hyperparameters}) and then applying GSLIB to generate one geological realization for each set of hyperparameters. We thus construct and simulate 2000 prior geomodels, which we denote as $\mathbf{m}_{i}$, $i=1, \ldots, 2000$ (other numbers of training samples will also be considered). Storage aquifer pressure and saturation fields at time $t=1$, 4, 7, 10, 13, 16, 20, 23, 26 and 30~years are saved. The networks are then trained (i.e., optimal network parameters are determined) such that the difference between predicted and simulated results is minimized. For detailed descriptions of the network architecture and training procedure, please refer to~\cite{tangmeng2022Deep, han2024surrogate}. The time required to train the pressure network (300~epochs) is about 10~hours, and that for the saturation network (500~epochs) is about 16~hours. These timings are with a single Nvidia Tesla V100 GPU (note that the A100 GPU, which is about twice as fast as the V100 GPU for training, could also be used).

We now quantify the prediction errors in the trained networks for pressure and saturation. A set of 500 new geomodels (referred to as the test set) is generated, using the same sampling procedure described above, and then simulated. The relative errors in pressure and saturation, $\delta_p^i$ and $\delta_S^i$, for each test-case sample $i$, are defined as follows:
\begin{equation} \label{error_p}
\delta_p^i = \frac{1}{n_sn_t} \sum_{j=1}^{n_s} \sum_{t=1}^{n_t} \frac{| (\hat{p}_s)_{i,j}^t - (p_s)_{i,j}^t |}{(p_s)_{i,\mathrm{max}}^t - (p_s)_{i,\mathrm{min}}^t},
\end{equation}
\begin{equation} \label{error_s}
\delta_S^i = \frac{1}{n_sn_t} \sum_{j=1}^{n_s} \sum_{t=1}^{n_t} \frac{| (\hat{S}_s)_{i,j}^t - (S_s)_{i,j}^t |}{(S_s)_{i,j}^t + \epsilon}, 
\end{equation}
where $i, j, t$ are the number index for test case, grid block, and time step, respectively, ${n_s}=43,200$ is the number of grid blocks in the storage aquifer, and $n_t=10$ is the number of time steps considered in the surrogate models. The quantities $\hat{p}_s$ and $\hat{S}_s$ are pressure and $\mathrm{CO}_{2}$ saturation predictions provided by the surrogate model, $p_s$ and $S_s$ are analogous predictions from ECLIPSE, $(p_s)_{i,\mathrm{max}}^t$ and $(p_s)_{i,\mathrm{min}}^t$ are the maximum and minimum pressure in test case $i$ at time step $t$, and $\epsilon$ = 0.025 is a constant included to avoid division by zero or near-zero values.

In Fig.~\ref{fig:relative_errors}, we present the relative errors for pressure and saturation (computed using Eqs.~\ref{error_p} and \ref{error_s}) for surrogate models trained with different numbers of runs. Results are shown as box plots, where the whiskers above and below the boxes indicate the $\mathrm{P}_{90}$ and $\mathrm{P}_{10}$ percentile errors, the top and bottom of each box show the $\mathrm{P}_{75}$ and $\mathrm{P}_{25}$ errors, and the orange line within the box is the $\mathrm{P}_{50}$ error. We see that all errors decrease with increasing numbers of training samples. With 2000 training runs, the $\mathrm{P}_{90}$ pressure and saturation relative errors are 0.014 and 0.037. Limited numerical experimentation with larger numbers of training samples did not provide consistent improvement in the $\mathrm{P}_{90}$ errors.

The errors at monitoring well locations are also of interest because these are the quantities actually used in data assimilation. These errors, for both pressure and saturation, were found to be unbiased (mean values very near zero). The $\mathrm{P}_{50}$ mean absolute errors for pressure and saturation are 0.05~MPa and 0.02 saturation units. These values are both smaller than the standard deviation of measurement error used in history matching (0.1~MPa and 0.05 saturation units). This, along with the unbiased nature of the surrogate model errors, suggest that the surrogate model is appropriate for use in data assimilation. The additional model error introduced by the surrogate model could be accounted for, if necessary, using the approach described in~\cite{han2024surrogate}.

\begin{figure}[!ht]
\centering
\begin{subfigure}[t]{.49\textwidth}
  \centering
  \resizebox{\linewidth}{!}{\input{Figures/surrogate_model/pressure_relative_errors}}
  \caption{Pressure relative errors}
  \label{fig:observed_CO2}
\end{subfigure}%
\hfill 
\begin{subfigure}[t]{.49\textwidth}
  \centering
  \resizebox{\linewidth}{!}{\input{Figures/surrogate_model/saturation_relative_errors}}
  \caption{Saturation relative errors}
  \label{fig:pressure}
\end{subfigure}
\caption{Pressure and saturation relative errors using the surrogate models trained with different number of training (ECLIPSE simulation) runs. Boxes show $\mathrm{P}_{90}$, $\mathrm{P}_{75}$, $\mathrm{P}_{50}$, $\mathrm{P}_{25}$ and $\mathrm{P}_{10}$ errors computed using Eqs.~\ref{error_p} and \ref{error_s}.}
\label{fig:relative_errors}
\end{figure}

\begin{figure}[H]
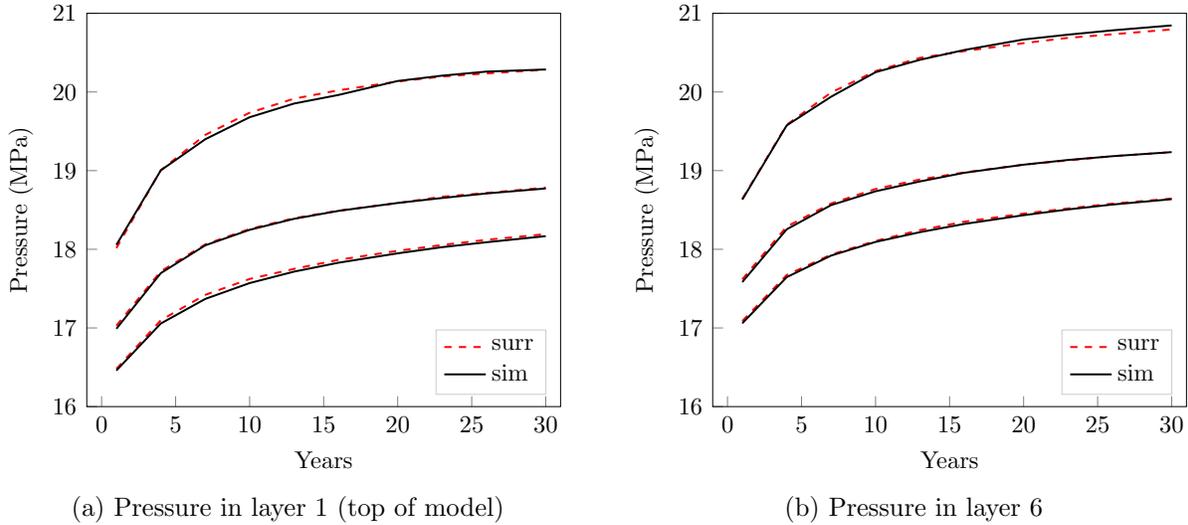

\centering
\begin{subfigure}[t]{.48\textwidth}
  \centering
  \resizebox{\linewidth}{!}{\input{Figures/surrogate_model/pressure_compare_0}}
  \caption{Pressure in layer~1 (top of model)}
  \label{fig:observed_CO2}
\end{subfigure}%
\hfill 
\begin{subfigure}[t]{.48\textwidth}
  \centering
  \resizebox{\linewidth}{!}{\input{Figures/surrogate_model/pressure_compare_6}}
  \caption{Pressure in layer~6}
  \label{fig:pressure}
\end{subfigure}
\caption{Comparison of pressure ensemble statistics from the surrogate model (red dashed curves) and simulation results (black solid curves) at grid blocks intersected by the monitoring well. The lower, middle and upper curves are $\mathrm{P}_{10}$, $\mathrm{P}_{50}$ and $\mathrm{P}_{90}$ results among the 500 test cases.}
\label{fig:surr_sim_pres}
\end{figure}

We next assess the performance of the surrogate model from two other perspectives. First, we compare the ensemble statistics for particular grid blocks intersected by the monitoring well. Results of this type are important because these quantities are used directly in the history matching procedure. Time-series for pressure, in layers~1 and 6 at the monitoring well location, are shown in Fig.~\ref{fig:surr_sim_pres}. Results are presented in terms of $\mathrm{P}_{10}$, $\mathrm{P}_{50}$ and $\mathrm{P}_{90}$ responses. Close agreement between the two sets of results is clearly observed, demonstrating the accuracy of the surrogate model for these local quantities.

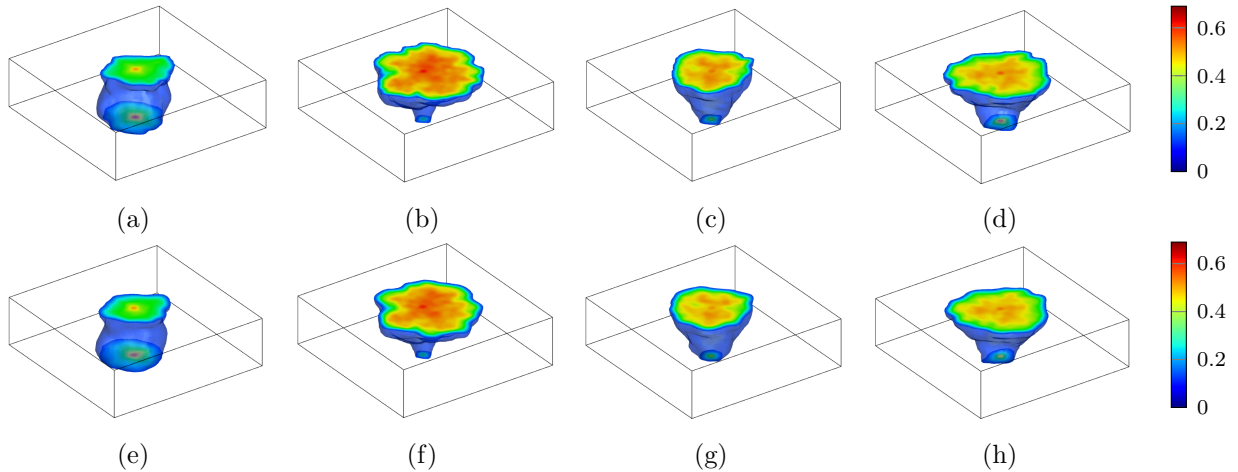
\begin{figure}[!h]
  \centering
  \subfloat[\centering]{\includegraphics[width=0.24\textwidth]{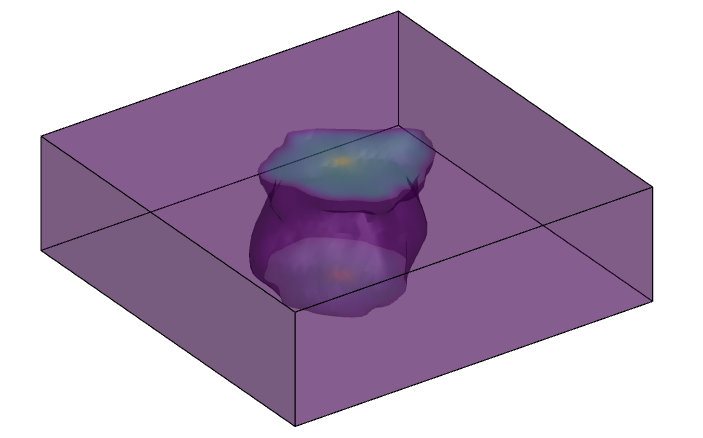}}
  \subfloat[\centering]{\includegraphics[width=0.24\textwidth]{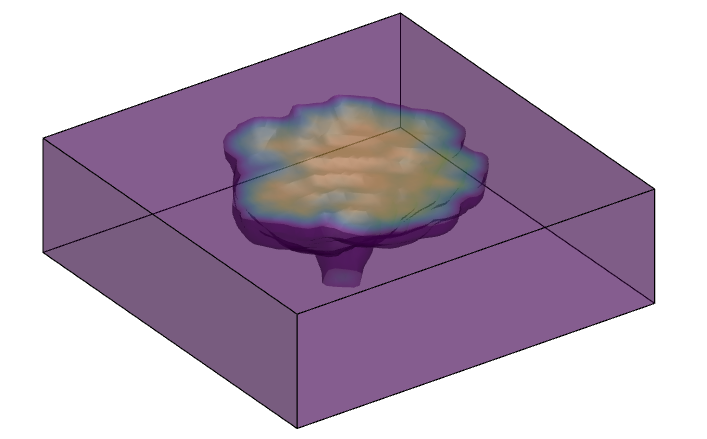}}
  \subfloat[\centering]{\includegraphics[width=0.24\textwidth]{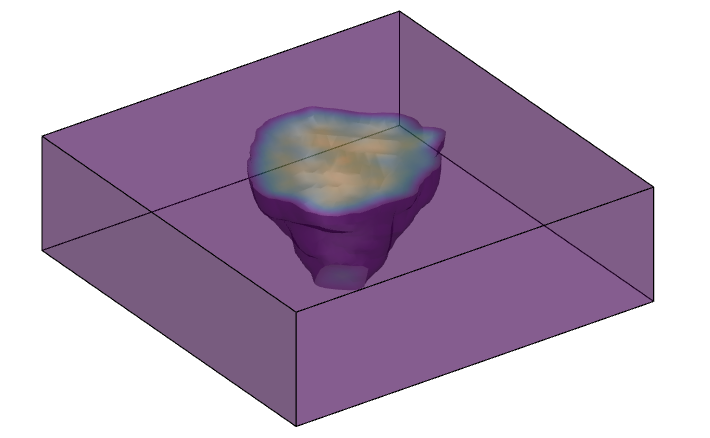}}
  \subfloat[\centering]{\includegraphics[width=0.24\textwidth]{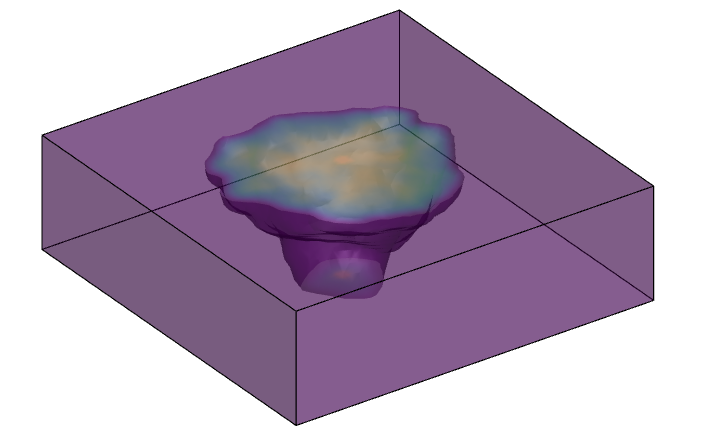}}
  \begin{tikzpicture}
    \begin{axis}[
      hide axis,
      scale only axis,
      height=2.2cm,
      width=0.02\textwidth,
      colorbar right,
      colorbar style={
        width=0.2cm, 
        font=\scriptsize,
      },
      colormap={customViridis}{
            rgb=(0.267, 0.004, 0.329)
            rgb=(0.282, 0.122, 0.439)
            rgb=(0.263, 0.247, 0.522)
            rgb=(0.212, 0.361, 0.553)
            rgb=(0.169, 0.459, 0.557)
            rgb=(0.129, 0.557, 0.553)
            rgb=(0.129, 0.651, 0.522)
            rgb=(0.251, 0.741, 0.447)
            rgb=(0.467, 0.820, 0.325)
            rgb=(0.729, 0.871, 0.157)
            rgb=(0.992, 0.906, 0.145)
      },
      colormap name=customViridis,
      point meta min=0.05,
      point meta max=0.6
    ]
      \addplot [draw=none] coordinates {(0,0)};
    \end{axis}
  \end{tikzpicture}
  \vspace{0.5 mm}
  \subfloat[\centering]{\includegraphics[width=0.24\textwidth]{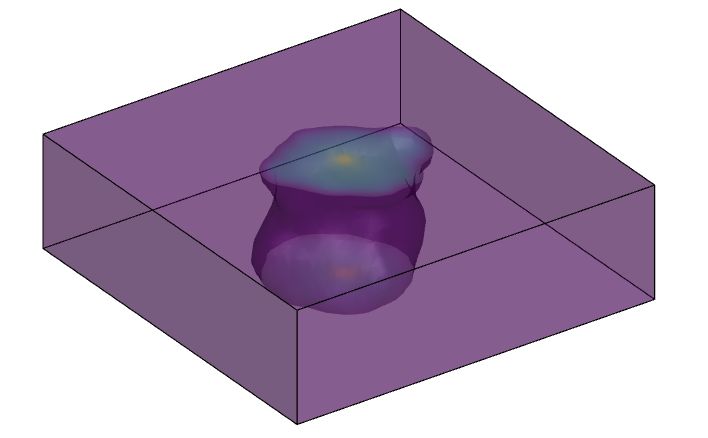}}
  \subfloat[\centering]{\includegraphics[width=0.24\textwidth]{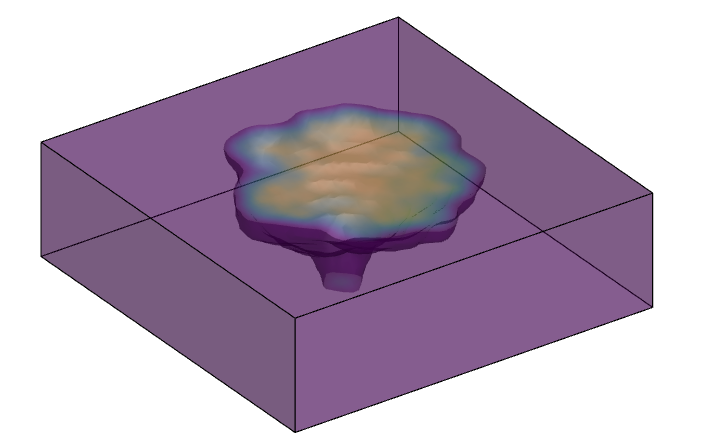}}
  \subfloat[\centering]{\includegraphics[width=0.24\textwidth]{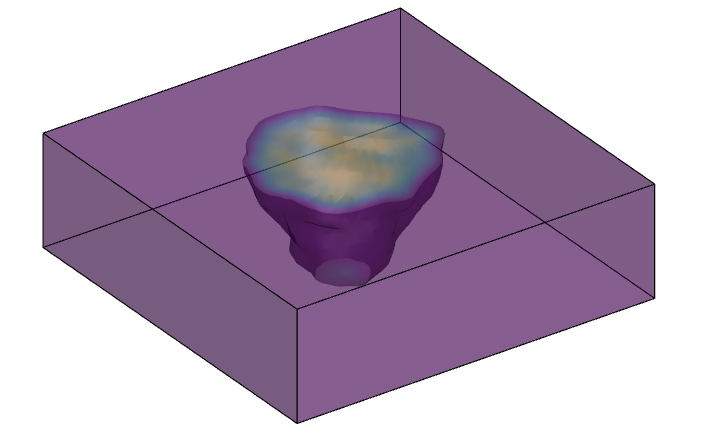}}
  \subfloat[\centering]{\includegraphics[width=0.24\textwidth]{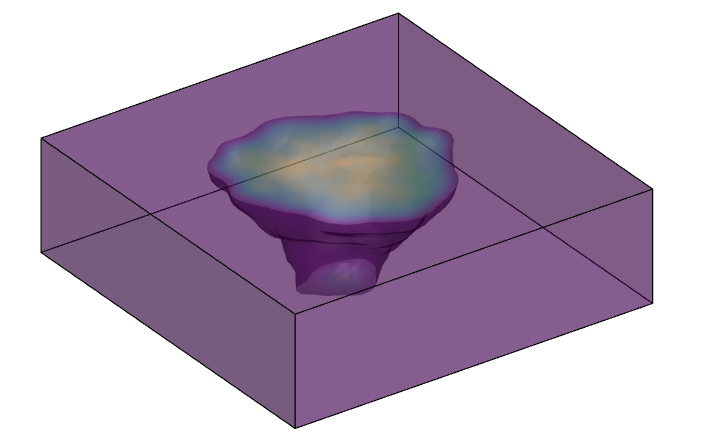}}
  \begin{tikzpicture}
    \begin{axis}[
      hide axis,
      scale only axis,
      height=2.2cm,
      width=0.02\textwidth,
      colorbar right,
      colorbar style={
        width=0.2cm,
        font=\scriptsize,
      },
      colormap={customViridis}{
            rgb=(0.267, 0.004, 0.329)
            rgb=(0.282, 0.122, 0.439)
            rgb=(0.263, 0.247, 0.522)
            rgb=(0.212, 0.361, 0.553)
            rgb=(0.169, 0.459, 0.557)
            rgb=(0.129, 0.557, 0.553)
            rgb=(0.129, 0.651, 0.522)
            rgb=(0.251, 0.741, 0.447)
            rgb=(0.467, 0.820, 0.325)
            rgb=(0.729, 0.871, 0.157)
            rgb=(0.992, 0.906, 0.145)
      },
      colormap name=customViridis,
      point meta min=0.05,
      point meta max=0.6
    ]
      \addplot [draw=none] coordinates {(0,0)};
    \end{axis}
  \end{tikzpicture}
  \vspace{0.5 mm}
  \subfloat[\centering]{\includegraphics[width=0.24\textwidth]{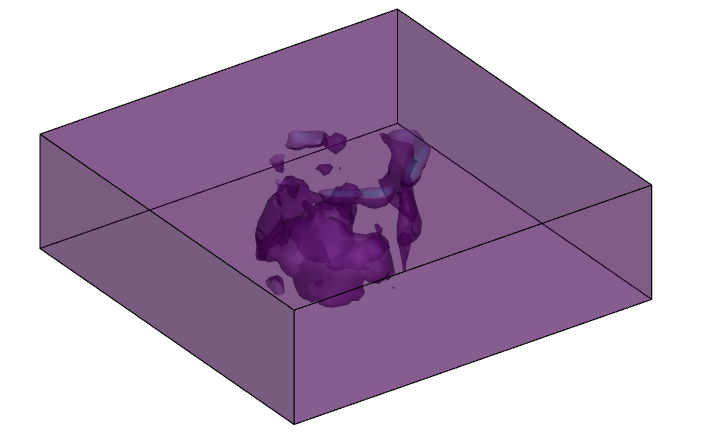}}
  \subfloat[\centering]{\includegraphics[width=0.24\textwidth]{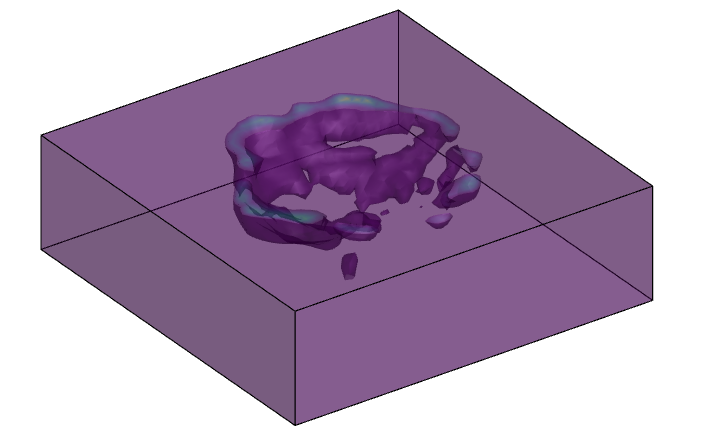}}
  \subfloat[\centering]{\includegraphics[width=0.24\textwidth]{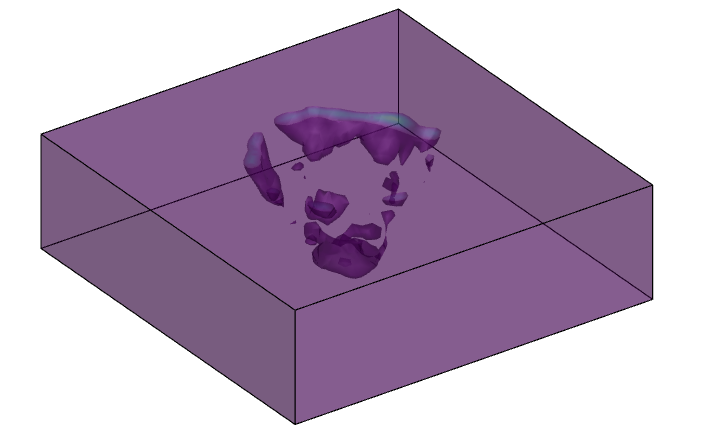}}
  \subfloat[\centering]{\includegraphics[width=0.24\textwidth]{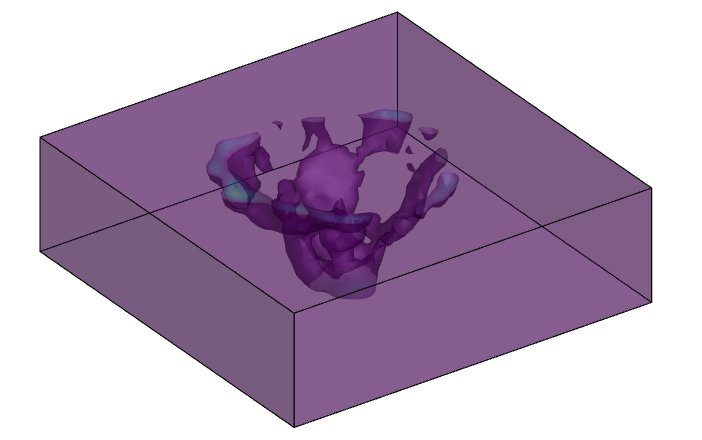}}
  \begin{tikzpicture}
    \begin{axis}[
      hide axis,
      scale only axis,
      height=2.2cm,
      width=0.02\textwidth,
      colorbar right,
      colorbar style={
        width=0.2cm,
        font=\scriptsize,
        tick label style={/pgf/number format/fixed}
      },
      colormap={customViridis}{
            rgb=(0.267, 0.004, 0.329)
            rgb=(0.282, 0.122, 0.439)
            rgb=(0.263, 0.247, 0.522)
            rgb=(0.212, 0.361, 0.553)
            rgb=(0.169, 0.459, 0.557)
            rgb=(0.129, 0.557, 0.553)
            rgb=(0.129, 0.651, 0.522)
            rgb=(0.251, 0.741, 0.447)
            rgb=(0.467, 0.820, 0.325)
            rgb=(0.729, 0.871, 0.157)
            rgb=(0.992, 0.906, 0.145)
      },
      colormap name=customViridis,
      point meta min=0.05,
      point meta max=0.3,
    ]
      \addplot [draw=none] coordinates {(0,0)};
    \end{axis}
  \end{tikzpicture}
  \caption{\( \text{CO}_{2} \) saturation at 30~years from ECLIPSE simulation (top row), the recurrent R-U-Net surrogate model (middle row), and their absolute difference (bottom row) for four representative realizations.}
  \label{fig:surr_sim_sat}
\end{figure}

We next compare field-scale solutions for $\mathrm{CO}_{2}$ saturation at the end of the injection period (30~years). We apply the $k$-means and $k$-medoids clustering procedure described in \cite{shirangi2016General, jiang2024History} to select distinct (representative) saturation fields. These fields are displayed in Fig.~\ref{fig:surr_sim_sat}. Simulation results are shown in the top row, surrogate predictions in the middle row, and their absolute differences in the bottom row (note the modified color scale in the difference plots). Different plume shapes and extents are evident between the four models. The surrogate predictions are visually close to the reference simulation results in all cases, consistent with the errors in Fig.~\ref{fig:relative_errors}(b). Error appears to be the largest around the saturation fronts. The saturation fields in the top layer of the model are seen to be quite accurate.

The surrogate model is trained with realizations characterized by hyperparameters sampled from their prior ranges. The synthetic true models used in this study are also (new) random samples from the prior. Thus, during history matching, the surrogate model is essentially performing a high-dimensional interpolation, since the data are (technically) contained within the prior. Of course, with 2000 (or even 10,000) realizations this is a relatively coarse sampling. The results presented in Figs.~\ref{fig:relative_errors}, \ref{fig:surr_sim_pres} and \ref{fig:surr_sim_sat} indicate that the sampling is, however, sufficient to enable the 3D recurrent R-U-Net surrogate to provide predictions of the accuracy required for history matching.

\section{Hierarchical Data Assimilation} \label{sec:hierarchical}

We begin this section with a description of the overall workflow for the hierarchical data assimilation procedure. The key algorithm used in our procedure, sequential Monte Carlo-based approximate Bayesian computation (SMC-ABC), is described in detail. Ensemble smoother with multiple data assimilation (ESMDA) is also discussed. Rejection sampling (RS), used as a reference, is then presented.

\subsection{Overall workflow}
Many traditional data assimilation procedures involve a fixed set of hyperparameters. In such cases, the grid-block permeabilities or porosities are typically the model parameters to be estimated. These high-dimensional discrete parameters can be inferred using Bayes' law:
\begin{equation} \label{traditional_Bayes}
  p(\mathbf{m} | \Bdobs) = \frac{p({\mathbf{m}}) p(\Bdobs| \mathbf{m})} {p(\Bdobs)},
\end{equation}
where $\mathbf{m}$ represents the discretized model parameters, $\Bdobs$ denotes the observed data, $p({\mathbf{m}})$ is the prior density, $p(\Bdobs| \mathbf{m})$ is the likelihood function, $p(\mathbf{m} | \Bdobs)$ is the posterior density, and $p(\Bdobs)$ is the evidence. The hyperparameters do not appear in this formulation because they are assumed to be known.

In hierarchical data assimilation, by contrast, the hyperparameters are treated as unknowns characterized by hyperprior distributions. This is a more appropriate treatment in real applications, since these parameters are very unlikely to be known with certainty. Bayes' rule for the joint posterior density over model parameters and hyperparameters, conditioned on the observed data, is now expressed as
\begin{equation} \label{Bayes}
  p(\mathbf{h}, \mathbf{m} | \Bdobs) = \frac{p(\bm{\mathrm{h}}, \bm{\mathrm{m}}) p(\bm{\mathrm{d}}_{\mathrm{obs}}|\bm{\mathrm{h}}, \bm{\mathrm{m}})} {p(\bm{\mathrm{d}}_{\mathrm{obs}})},
\end{equation}
where $\mathbf{h}$ is the set of hyperparameters, $p(\mathbf{h}, \mathbf{m})$ is the joint prior density, and $p(\mathbf{h}, \mathbf{m} | \Bdobs)$ is the joint posterior density. Here $\mathbf{h}=[\mu_{\log k}, \sigma_{\log k}, \log_{10}a_r, l_h, \phi]$, as described in Section~\ref{sec:model setup}. 
In this hierarchical Bayesian inference problem, the joint parameter space is very high-dimensional. We therefore decompose this joint space into two parts as follows:
\begin{equation} \label{Bayes}
  p(\mathbf{h}, \mathbf{m} | \Bdobs) = p(\mathbf{h} | \Bdobs)
  p(\mathbf{m} | \mathbf{h},\Bdobs).
\end{equation}
Here $p(\mathbf{h} | \Bdobs)$ is the marginal posterior density of the hyperparameters and $p(\mathbf{m} | \mathbf{h},\Bdobs)$ is the conditional posterior density of the discretized model parameters. The procedures applied in this work enable sampling from these two densities, which allows us to explore the joint posterior distribution. We now describe the approaches used for these samplings.

The marginal posterior density of the hyperparameters, conditioned on the measurements, is again given by Bayes' theorem, in this case expressed as
\begin{equation} \label{marginal}
  p(\bm{\mathrm{h}} | \bm{\mathrm{d}}_{\mathrm{obs}}) = \frac{p(\bm{\mathrm{h}}) p(\Bdobs| \bm{\mathrm{h}})} {p(\Bdobs)}.
\end{equation}
Here $p(\mathbf{h})$ is the prior density of the hyperparameters and $p(\Bdobs | \mathbf{h})$ is the likelihood function, which is given by
\begin{equation} \label{likelihood}
  p(\bm{\mathrm{d}}_{\mathrm{obs}}| \bm{\mathrm{h}}) = \int p(\bm{\mathrm{d}}_{\mathrm{obs}}|\bm{\mathrm{m}}, \bm{\mathrm{h}}) p(\bm{\mathrm{m}} | \bm{\mathrm{h}})d\bm{\mathrm{m}}.
\end{equation}
This likelihood is computationally intractable, as it involves an integral over all possible model parameters. We thus apply the likelihood-free (or simulation-based) inference procedure, specifically the SMC-ABC method, to draw hyperparameter samples from the marginal posterior distribution. The conditional posterior density of discrete model parameters, $p(\mathbf{m} | \mathbf{h},\Bdobs)$, is provided by traditional data assimilation methods with a fixed set of hyperparameters. In this study, we apply the widely used ESMDA procedure for this step, i.e., to estimate grid-block permeability values for multiple realizations, with a specified set of hyperparameters.

The overall workflow for the hierarchical methodology is shown in Algorithm~\ref{hierarchical}. As indicated there, we first draw posterior samples from the marginal posterior distribution of hyperparameters. The representative samples are then selected from the posterior hyperparameters. For each set of hyperparameters, we estimate grid-block permeability conditioned on hyperparameters and observed data. The detailed procedures for these steps are described in the next sections.

\begin{algorithm}
\setstretch{1.5}
\caption{Hierarchical data assimilation} \label{hierarchical}
\begin{algorithmic}[1]
\STATE Draw posterior samples from the marginal posterior distribution of hyperparameters $p(\mathbf{h} | \Bdobs)$.
\STATE Select $N$ representative samples from the posterior hyperparameters.
\FOR{$i = 1$ to $N$}
  \STATE Estimate posterior permeability field conditioned on hyperparameters and observations.
  \STATE Obtain joint posterior samples of hyperparameters and permeability realizations.
\ENDFOR
\end{algorithmic}
\end{algorithm}

\subsection{Sequential Monte Carlo-based approximate Bayesian computation method}
\label{sec:smc-abc}

The SMC-ABC method is used to estimate the marginal posterior distribution of the hyperparameters. In the hierarchical data assimilation examples, we consider $\mathbf{h}=[\mu_{\log k}, \sigma_{\log k}, \log_{10}a_r]$, which is a subset of the hyperparameters considered for surrogate model training. This is because the monitoring data in our setup were found to be much less sensitive to the other two hyperparameters, $l_h$ and $\phi$. Thus these quantities are fixed in our numerical examples, which provides computational savings.

SMC-ABC is a likelihood-free (or simulation-based) inference algorithm to deal with computationally intractable likelihood functions in Bayesian inference problems. The standard ABC method simply accepts parameter samples that generate simulated data resembling the observations. SMC-ABC is a more efficient approximate inference method. Based on the idea of importance sampling, it focuses computational effort on regions of the parameter space that are more likely to produce a close match between observed and simulated data. The sequential sampling strategy essentially moves a population of particles or samples (sets of hyperparameters) toward the target posterior by constructing a sequence of intermediate distributions.

The details of the SMC-ABC method are provided in Algorithm~\ref{smc_abc}. The procedure applies a decreasing sequence of thresholds, where the threshold is based on distance values. The distance function, given below, quantifies the discrepancy between the observed and simulated data. Particles are accepted if their distance is below the threshold; otherwise they are rejected. The thresholds can be specified in advance or determined adaptively at each iteration. The latter approach is used here. In the first iteration, the threshold is set to infinity, which means all particles are accepted. In subsequent iterations, the threshold is taken to be the median of the accepted distances from the previous iteration. The number of accepted hyperparameter samples (or population size) remains constant throughout all iterations of the SMC-ABC algorithm.

There are many possible forms for the distance function in ABC-based methods. In this study, we use the weighted Euclidean distance function between the simulated and observed data. Recall that in our case these data are time-series of pressure and saturation at the monitoring well. The distance/mismatch function used in this work is given by 
\begin{equation} \label{distance}
d_m(\mathbf{y}, \mathbf{d})= \sum_{i=1}^n\left(\frac{\mathbf{y}_i^p - \mathbf{d}_i^p}{\sigma_i^p}\right)^2 + \sum_{i=1}^n\left(\frac{\mathbf{y}_i^s - \mathbf{d}_i^s}{\sigma_i^s}\right)^2,
\end{equation}
where $\mathbf{y}$ represents the simulated data, $\mathbf{d}$ indicates the observed data, $n$ is the number of time steps in the historical period, $p$ and $s$ denote pressure and saturation, and $\sigma^p$ and $\sigma^s$ (here taken to be the standard deviations of measurement error) scale the mismatch terms. The scaling could be adaptively updated at each iteration~\cite{prangle2017adapting}, though this was not explored here.

In the first iteration of the SMC-ABC algorithm the hyperparameter samples are drawn from the prior distribution. Geological realizations are then generated using sequential Gaussian simulation. The simulated data are obtained using the surrogate model and the distance to the observed data is calculated using Eq.~\ref{distance}. All samples in the first iteration are accepted and weighted equally. Weighting is applied in subsequent iterations. We draw hyperparameter samples from the previous weighted population and then perturb the samples using a perturbation kernel.

Specifically, the hyperparameter samples in each iteration are drawn from the proposal distribution given by

\begin{equation} \label{proposal}
q_t(\mathbf{h})= 
\begin{cases}
p(\mathbf{h}) & \text{if } t=1, \\ 
\displaystyle\sum_{j=1}^N w_{t-1}^{(j)} K_t\left(\mathbf{h} | \mathbf{h}_{t-1}^{(j)}\right) & \text{if } t>1,
\end{cases}
\end{equation}
where $t$ is the iteration index, $N$ is the number of particles (or population size), $w$ is the importance weight, and $K_t$ is the perturbation kernel. As suggested in \cite{beaumont2009adaptive}, here we use a multivariate Gaussian distribution as the kernel:
\begin{equation} \label{kernel}
K_t\left(\mathbf{h} | \mathbf{h}_{t-1}^{(j)}\right)=\mathcal{N}\left(\mathbf{h}_{t-1}^{(j)}, 2 \mathbf{\Sigma}_{t-1}\right),
\end{equation}
where $\mathbf{\Sigma}_{t-1}$ is the weighted empirical covariance of the accepted hyperparameter samples in the previous iteration. A hyperparameter sample is accepted if its distance (to the observed data) is lower than the threshold. The weight assigned to each accepted particle is given by
\begin{equation} \label{weight}
w_t^{(i)} = \frac{p(\mathbf{h}_t^{(i)})}{q_t(\mathbf{h}_t^{(i)})}.
\end{equation}
These weights then impact the proposal distribution through Eq.~\ref{proposal}. For a detailed derivation of the SMC-ABC algorithm used in this study, please refer to~\cite{toni2009Approximate, beaumont2009adaptive}.

The above procedure is repeated for a sequence of iterations. With a large number of iterations, the distribution will better approximate the target posterior distribution. In practice, the algorithm requires a termination criterion. In this study, the algorithm is terminated once the acceptance rate is less than 5\%. This typically corresponds to 7--9 iterations in SMC-ABC for our problem. Alternative stopping rules could be based on a predefined number of iterations, a fixed computational budget, or the stabilization of posterior approximations.

\begin{algorithm}[H]
\setstretch{1.5}
\caption{SMC-ABC procedure} \label{smc_abc}
\begin{algorithmic}[1]
\STATE Set the iteration counter $t = 1$, initial distance threshold $\epsilon_1 = \infty$, and the population size $N$.
\REPEAT
    \STATE Set the sample (or particle) indicator $i = 1$.
    \REPEAT
       \STATE Draw hyperparameter sample {\bf h} from the proposal distribution in Eq.~\ref{proposal}.
       \STATE Reject and return to (5), if {$p(\bf h) = 0$}.
       \STATE Generate geological realization using sequential Gaussian simulation.
       \STATE Predict pressure and saturation using the surrogate model.
       \STATE Calculate distance $d_m$ between simulated and observed data using Eq.~\ref{distance}.
       \STATE Accept the hyperparameter sample if $d_m \leq \epsilon_t$ and store the distance as $(d_m)_t^i$.
       \STATE Increment $i$ to $i + 1$.
    \UNTIL $N$ hyperparameter samples are accepted.
\STATE Update the importance weights $w_t^i$ of the accepted particles using Eq.~\ref{weight}.
\STATE Set the threshold $\epsilon_{t+1}$ as the median of the $(d_m)_t^i$ values.
\STATE Increment $t$ to $t + 1$.
\UNTIL Termination criterion is reached.
\end{algorithmic}
\end{algorithm}
\subsection{Ensemble smoother with multiple data assimilation procedure}
\label{sec:esmda}

As noted earlier, we apply ESMDA to estimate the grid-block permeabilities, for particular sets of hyperparameters {\bf h}, given time-series monitoring-well data. In this study, the particular sets of hyperparameters are selected from the weighted posterior samples obtained using SMC-ABC. Specifically, we first apply systematic resampling~\cite{kitagawa1996Monte, arulampalam2002Tutorial} to obtain equally weighted posterior samples. The $k$-means and $k$-medoids procedure~\cite{shirangi2016General} is then used to select 10 sets of representative hyperparameter samples.

Ensemble-based methods can be divided into two categories: filter and smoother. Filter methods assimilate sequentially in time, while smoother methods assimilate all observations simultaneously. ESMDA, which is an iterative form of ES introduced by \citet{emerick2013ensemble}, is widely used in reservoir engineering applications. It is a derivative-free method that applies statistical linearization using an ensemble. The full covariance matrix is approximated by a low-rank ensemble-based representation, so computational costs are manageable. ESMDA is straightforward to implement and parallelizes naturally.

In ESMDA, the same observed data are assimilated multiple times with an inflated measurement error. The objective function in ESMDA is formulated as a regularized minimization problem, defined as follows:
\begin{equation} \label{objective}
J(\bm{\mathrm{m}}) = \frac{1}{2} \big|\Bdobs-f(\bm{\mathrm{m}}) \big|_{\bm{\mathrm{R}}}^2 + \frac{1}{2} \big|\bm{\mathrm{m}}-\bar{\bm{\mathrm{m}}} \big|_{\bm{\mathrm{\Sigma}}}^2,
\end{equation}
where $\bar{\mathbf{m}}$ is the prior mean of model parameters, $\mathbf{R}$ is measurement error covariance, $\mathbf{\Sigma}$ represents the prior covariance, and $f$ denotes function evaluation via the surrogate model. Minimizing this objection function is equivalent to the maximum a posteriori estimator from a Bayesian perspective.

The details of the ESMDA procedure are given in Algorithm~\ref{esmda}. The analysis or update equation in ESMDA is 
\begin{equation}\label{es_update}
\mathbf{m}_{i}^a=\mathbf{m}_i^f+\mathbf{C}_{md}^f\left(\mathbf{C}_{dd}^f+\alpha \mathbf{R}\right)^{-1}\left[\mathbf{d}_{\mathrm{uc}, i} - f(\mathbf{m}_i^f)\right],
\end{equation}
for $i = 1, \ldots, N_e$, where $N_e$ is the ensemble size, $\mathbf{m}^a$ and $\mathbf{m}^f$ represent the analyzed and forecasted parameter fields, respectively, $\mathbf{C}_{md}^f$ is the cross-covariance between model parameters and predicted data, $\mathbf{C}_{dd}^f$ is the auto-covariance of predicted data, $\alpha$ represents the inflation coefficient, and $\mathbf{d}_\mathrm{uc}$ contains the perturbed observations. These are given by
\begin{equation}\label{perturb}
\mathbf{d}_\mathrm{uc}=\Bdobs + \sqrt{\alpha} \mathbf{R}^{1/2} \textbf{z},
\end{equation}
where $\textbf{z} \sim \mathcal{N}(\textbf{0}, \textbf{I})$. The inflation coefficients $\alpha_j$ are selected to satisfy the constraint $\sum_{j=1}^{N_a}\alpha_j^{-1} = 1$, where $N_a$ is the number of data assimilation steps. A nonascending set of $\alpha_j$ is typically used.

\begin{algorithm}
\setstretch{1.5}
\caption{ESMDA procedure} \label{esmda}
\begin{algorithmic}[1]
\STATE Specify the number of data assimilation steps $N_a$, and the corresponding inflation coefficients $\alpha_j$, $j = 1, \ldots, N_a$.
\STATE Construct the initial ensemble of realizations.
\FOR{$j = 1$ to $N_a$}
    \STATE Evaluate the ensemble using the surrogate model.
    \STATE Generate perturbed measurements $\mathbf{d}_\mathrm{uc, i}^j$ using Eq.~\eqref{perturb} for each ensemble member.
    \STATE Update the ensemble members using Eq.~\eqref{es_update}.
\ENDFOR
\end{algorithmic}
\end{algorithm}

As described earlier, we apply systematic resampling to obtain equally weighted samples and then use the $k$-means and $k$-medoids procedure to select 10 representative samples of posterior hyperparameters. Thus, 10 full ESMDA runs are required to estimate permeability fields given the hyperparameters and observed data. For each ESMDA run, we set $N_e$ = 500, $N_a$ = 4, and $\alpha_1$ = 9.333, $\alpha_2$ = 7.0, $\alpha_3$ = 4.0, $\alpha_4 = 2.0$. Therefore, in this stage we require a total of $500\times4\times10=20,000$ function evaluations.

\subsection{Rejection sampling}

RS is a rigorous sampling method that can generate independent posterior samples for the marginal posterior distribution of the hyperparameters. RS is used to provide reference results in this study. However, due to its very low acceptance rate, RS may require millions of forward function evaluations. The deep learning-based surrogate model makes the use of this algorithm feasible for our application, though RS is not expected to be suitable for more general settings.

RS is a Monte Carlo scheme that provides samples from the target distribution~\cite{ripley2009stochastic}. The basic idea of RS is to sample from a proposal distribution and then either accept or reject the proposed sample. The details of the RS algorithm are provided in Algorithm~\ref{RS}. We first draw a hyperparameter sample from the prior distribution and construct a geomodel realization using sequential Gaussian simulation. We then use the surrogate model to perform flow simulation and calculate the corresponding likelihood value. A sample $u$ is obtained from a uniform distribution over $[0,1]$. The prior sample is accepted or rejected based on the criterion
\begin{equation} \label{acceptance_criteria}
u \leq \frac{p(\Bdobs|\mathbf{h}, \mathbf{m})}{S_L},
\end{equation}
where $S_L$ is a bound constant. The estimation of a tight bound for \(S_L\) is important to the performance of RS, as a smaller value of \(S_L\) can lead to a larger acceptance rate. In this work, we set \(S_L\) as the maximum likelihood value of all prior samples. This likelihood can be estimated by
\begin{equation} \label{rejection_likelihood}
p(\Bdobs|\mathbf{h}, \mathbf{m})=\operatorname{det}\left(2 \pi \mathbf{R}\right)^{-\frac{1}{2}} \exp \left(-\frac{1}{2}(\Bdobs-f(\mathbf{m}))^T \boldsymbol{\mathbf{R}}^{-1}(\Bdobs-f(\mathbf{m}))\right).
\end{equation}

In this study, RS is applied as a reference to obtain stabilized marginal posterior distributions of the hyperparameters. The RS runs require ${\mathcal O}(10^6-10^7)$ function evaluations, though only ${\mathcal O}(10^2-10^3)$ samples are accepted. This corresponds to acceptance rates of $\sim10^{-4}$ for our problem.

\begin{algorithm}[H]
\setstretch{1.5}
\caption{Rejection sampling} \label{RS}
\begin{algorithmic}[1]
\STATE Sample a set of hyperparameters $\mathbf{h}$ from the prior distribution.
\STATE Use geostatistical simulation to generate one realization for the hyperparameters.
\STATE Run the surrogate model and compute the likelihood value using Eq.~\eqref{rejection_likelihood}.
\STATE Obtain $u$ from a uniform distribution over $[0,1]$.
\STATE Accept $\mathbf{h}$ if Eq.~\eqref{acceptance_criteria} is satisfied. Otherwise, reject $\mathbf{h}$ and return to step 1.
\end{algorithmic}
\end{algorithm}
\section{Data Assimilation Results}
\label{sec:results}

In this section, we apply the hierarchical procedure for two synthetic ‘true’ models. We first describe the problem setup. Posterior results are then generated and compared to those from the (reference) RS method. We quantify the substantial computational gains achieved using the hierarchical data assimilation approach.

\subsection{Problem setup for data assimilation}

To enable comparison with the rigorous RS, a limited number of observations is considered (otherwise the RS acceptance rate would be too low to provide meaningful results). Ensemble-based methods~\cite{evensen2004Sampling, emerick2016Analysis} can be used when large amounts of data are to be assimilated, though the posterior results may then be only approximate, as we will see in Section~\ref{sec:Traditional Data Assimilation}. The pressure and saturation data used in our history matching runs are measured in the top layer of the storage aquifer through one monitoring well, which is located near the injector (see Fig.~\ref{fig:well_location}). The assimilated data are at times of 1, 4, 7, 10, and 13~years (consistent with surrogate model output). The number of total measurements is thus 10. We consider Gaussian observational noise, of standard deviation 0.1~MPa for pressure, and 0.05 saturation units for saturation.

The synthetic true models are constructed by first randomly sampling the hyperparameters from the hyperprior distributions, and then generating a realization based on these hyperparameters using GSLIB. The values of the hyperparameters for the two true models, along with the prior ranges, are given in Table~\ref{table:true_models}. As mentioned in Section~\ref{sec:smc-abc}, the horizontal correlation length and porosity are fixed in the data assimilation procedure. The porosity is 0.2 for both true models, and the horizontal correlation length is 1200~m for true model~1 and 1920~m for true model~2. The synthetic observed data are generated using surrogate model predictions for the true model (true data) combined with the additive Gaussian measurement errors.

\begin{figure}[!ht]
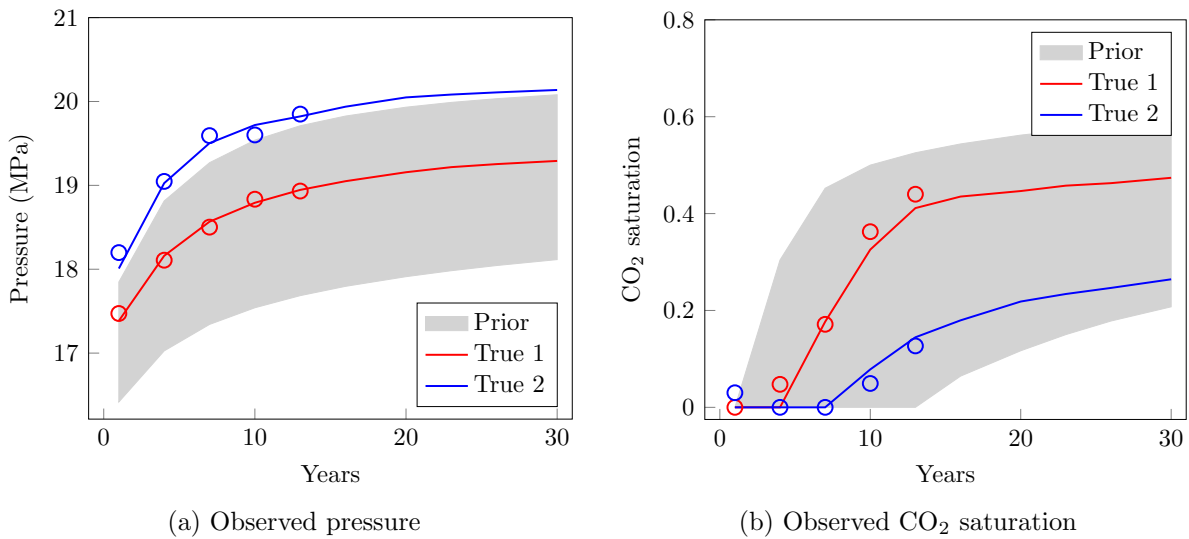

\centering
\begin{subfigure}[t]{.49\textwidth}
  \centering
  \resizebox{\linewidth}{!}{\input{Figures/setup_assimilation/pressure_data}}
  \caption{Observed pressure}
  \label{fig:pressure}
\end{subfigure}
\hfill
\begin{subfigure}[t]{.49\textwidth}
  \centering
  \resizebox{\linewidth}{!}{\input{Figures/setup_assimilation/saturation_data}}
  \caption{Observed \( \text{CO}_{2} \) saturation}
  \label{fig:observed_CO2}
\end{subfigure}%
\caption{Observed data (circles) and true responses (solid curves) at the monitoring location for two synthetic true models. Gray regions denote the $\mathrm{P}_{10}$--$\mathrm{P}_{90}$ range of the prior data for true model~2 (which differs slightly from the true model~1 prior range due to differences in $l_h$).}
\label{fig:observed_data}
\end{figure}

The observed pressure and saturation at the monitoring location for the two synthetic true models are shown in Fig.~\ref{fig:observed_data}. The red and blue curves represent the true model responses, and the circles indicate the observed data (which deviate from the curves due to noise). True model~2 shows higher pressure and lower saturation compared to true model~1. This difference is due to the lower values for both the mean of log-permeability ($\mu_{\log k}$) and permeability anisotropy ratio ($\log_{10}a_r$) in true model~2.

The gray regions in Fig.~\ref{fig:observed_data} represent the $\mathrm{P}_{10}$--$\mathrm{P}_{90}$ range of the prior model predictions for true model~2. The two synthetic models have almost the same prior prediction ranges, though there are slight differences due to the different correlation lengths used for the two true models. Note that the observed pressure data for true model~2 falls beyond the $\mathrm{P}_{90}$ of its prior distribution. The fact that these data fall near the edge of the distribution can pose challenges for data assimilation and parameter estimation.

\begin{table}[!ht]
\small
\setstretch{1.5}
\centering
\caption{Hyperprior distributions and hyperparameter values for two synthetic true models}
\label{table:true_models}
\begin{tabular}{p{4.2cm} p{3.1cm} p{3.1cm} p{3.1cm}}
\hline
\noalign{\hrule height 0.8pt}
 & $\mu_{\log k}$ & $\sigma_{\log k}$ & $\log_{10}a_r$ \\ 
\hline
Prior distribution  & $U$(2.5, 4.5) & $U$(0.5, 2) & $U$(-2, 0) \\ 
True model 1 & 3.3 & 0.9 & -0.5 \\ 
True model 2 & 2.7 & 1.2 & -1.7 \\ 
\hline
\noalign{\hrule height 0.8pt}
\end{tabular}
\end{table}

\subsection{Data assimilation results for true model~1}
\label{sec:true model 1}

\begin{figure}[!ht]
\centering
\begin{subfigure}[t]{.33\textwidth}
  \centering
  \resizebox{\linewidth}{!}{\includegraphics{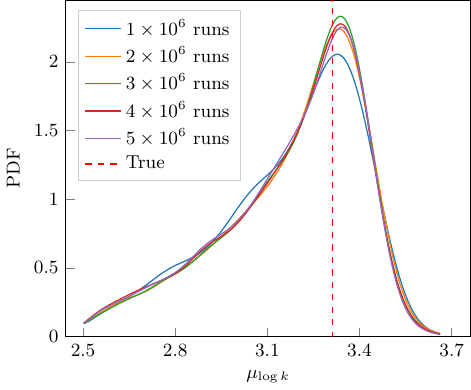}}
  \caption{$\mu_{\log k}$}
  \label{fig:RS5_perm_mean}
\end{subfigure}%
\hfill
\begin{subfigure}[t]{.33\textwidth}
  \centering
  \resizebox{\linewidth}{!}{\includegraphics{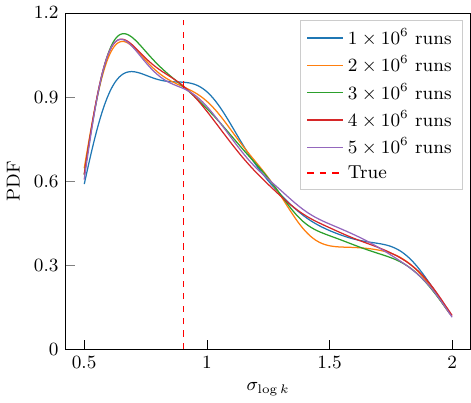}}
  \caption{$\sigma_{\log k}$}
  \label{fig:RS5_perm_std}
\end{subfigure}
\hfill 
\begin{subfigure}[t]{.33\textwidth}
  \centering
  \resizebox{\linewidth}{!}{\includegraphics{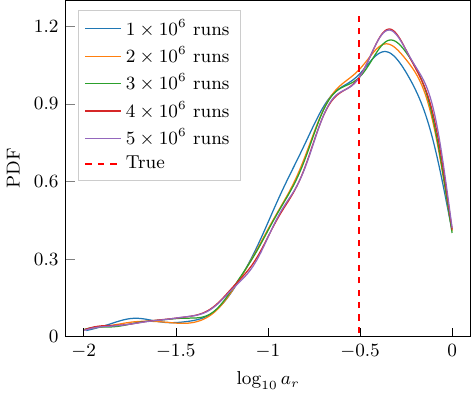}}
  \caption{$\log_{10}a_r$}
  \label{fig:RS5_kzkx}
\end{subfigure}
\caption{Marginal posterior distributions of hyperparameters obtained using RS for true model~1. Solid curves represent the posterior distributions obtained using different numbers of forward runs. Dashed vertical lines denote the true values of the hyperparameters.}
\label{fig:reference_marginal}
\end{figure}

Results for the posterior distributions of the hyperparameters obtained using RS are shown in Fig.~\ref{fig:reference_marginal}. The solid curves of various colors represent the posterior distributions generated using different numbers of forward runs, and the dashed red vertical lines denote the true values of the hyperparameters. Note that each forward run requires both a geostatistical simulation (to generate the geomodel) and a surrogate run. It is evident that, as the number of function evaluations increases, the RS posterior distributions appear to ‘converge.’ This behavior will be quantified below when we compute the Jensen–Shannon divergence for the posterior distributions.

We see that the modes of the RS posterior distributions do not precisely correspond to the true hyperparameter values. In idealized scenarios, such as a linear Gaussian system with noise-free observations and a prior centered at the true parameter value, the posterior mode is guaranteed to align with the true value. Several factors in our problem, however, including measurement noise, a nonlinear forward model, and the choice of prior, can act to shift the posterior modes away from the true values~\cite{tarantolaInverseProblemTheory2005b, stuartInverseProblemsBayesian2010}. In addition, the results in Fig.~\ref{fig:reference_marginal} correspond to the marginal posterior distributions, and their individual modes do not necessarily coincide with the mode of the joint posterior distribution.

In the comparisons that follow, we use the RS results derived from $4\times10^6$ runs as the reference results for true model~1. In this case, 2584 samples are accepted. The deep-learning-based surrogate model is clearly essential for this assessment, given the need for millions of function evaluations.

We next assess the performance of the SMC-ABC method for this problem. For an appropriate comparison to the reference RS results, we set the number of accepted samples (population size) in the SMC-ABC iterations to be consistent with the number of samples accepted in the converged RS procedure. For true model~1, the population size in SMC-ABC is set to 2500, which is close to the number of accepted samples (2584) in the reference RS results. This treatment ensures that the SMC-ABC and reference RS results have the same level of Monte Carlo approximation to the true posterior distributions.

The accuracy of the estimated posterior distributions of hyperparameters is quantified in terms of the Jensen--Shannon (JS) divergence. JS divergence is a statistical distance metric that measures the difference between two distributions, which in our case are the reference RS and approximate SMC-ABC results. JS divergence is defined as
\begin{equation} \label{jsd}
\mathrm{JS}(P \Vert Q)=\frac{1}{2} \mathrm{KL}(P \Vert M)+\frac{1}{2} \mathrm{KL}(Q \Vert M),
\end{equation}
where $P$ represents the approximate marginal posterior distribution of hyperparameters, $Q$ denotes the ‘exact’ marginal posterior distribution (taken to be the RS result with $4\times10^6$ runs), and $M=\frac{1}{2}(P+Q)$ is a mixture distribution of $P$ and $Q$. The Kullback--Leibler (KL) divergence in Eq.~\ref{jsd} is given by
\begin{equation} \label{kld}
\mathrm{KL}(P \Vert M)  =\int P(h) \log\Bigl(\frac{P(h)}{M(h)}\Bigr)\,\mathrm{d}h.
\end{equation}
Thus we see that the JS divergence is a symmetrized and smoothed version of the KL divergence.

\begin{figure}[!ht]
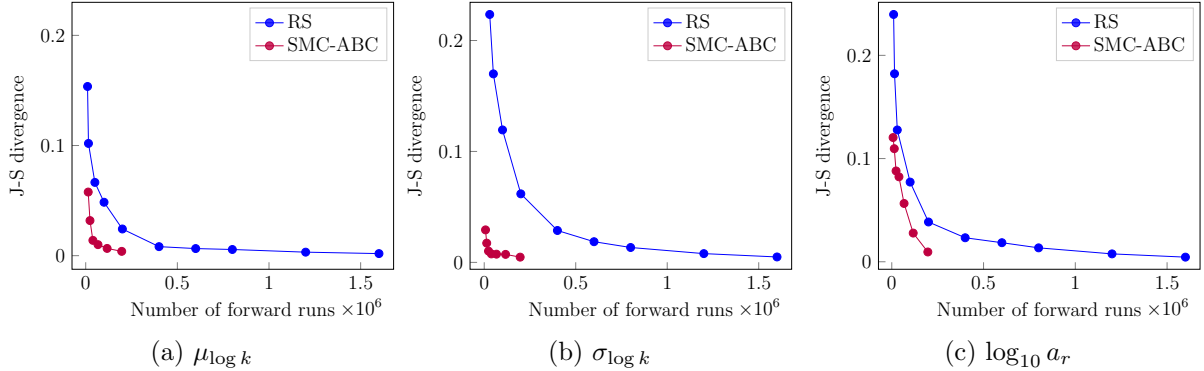

\centering
\begin{subfigure}[t]{.33\textwidth}
  \centering
  \resizebox{\linewidth}{!}{\input{Figures/efficiency_comparison/Evals_mean}}
  \caption{$\mu_{\log k}$}
  \label{fig:subplot_a}
\end{subfigure}%
\hfill 
\begin{subfigure}[t]{.33\textwidth}
  \centering
  \resizebox{\linewidth}{!}{\input{Figures/efficiency_comparison/Evals_std}}
  \caption{$\sigma_{\log k}$}
  \label{fig:subplot_b}
\end{subfigure}
\hfill 
\begin{subfigure}[t]{.33\textwidth}
  \centering
  \resizebox{\linewidth}{!}{\input{Figures/efficiency_comparison/Evals_kzkx}}
  \caption{$\log_{10}a_r$}
  \label{fig:subplot_c}
\end{subfigure}
\caption{Comparison of the computational efficiency for hyperparameter estimation using the RS and SMC-ABC methods. A smaller JS divergence means a more accurate estimation of the posterior distribution.}
\label{fig:efficiency_comparison}
\end{figure}

Results demonstrating the performance of the SMC-ABC method, with different numbers of forward runs, are presented in Fig.~\ref{fig:efficiency_comparison}. With the number of SMC-ABC iterations increasing, the distance threshold in each iteration becomes smaller, which leads to more function evaluations. We also show the behavior of RS in terms of this metric. Immediately apparent are the large differences between the RS curves (in blue) and the SMC-ABC curves (in red). RS displays large JS divergence relative to the reference, and fairly slow convergence. The opposite is observed for SMC-ABC. This difference is particularly evident in the results for $\sigma_{\log k}$ (Fig.~\ref{fig:efficiency_comparison}(b)). These results show that around $0.2\times10^6$ forward runs using SMC-ABC result in a near-zero JS divergence, indicating close agreement with the reference RS results. Therefore, we use the SMC-ABC results derived from $0.2\times10^6$ forward runs to compare with the reference RS.

\begin{figure}[!ht]
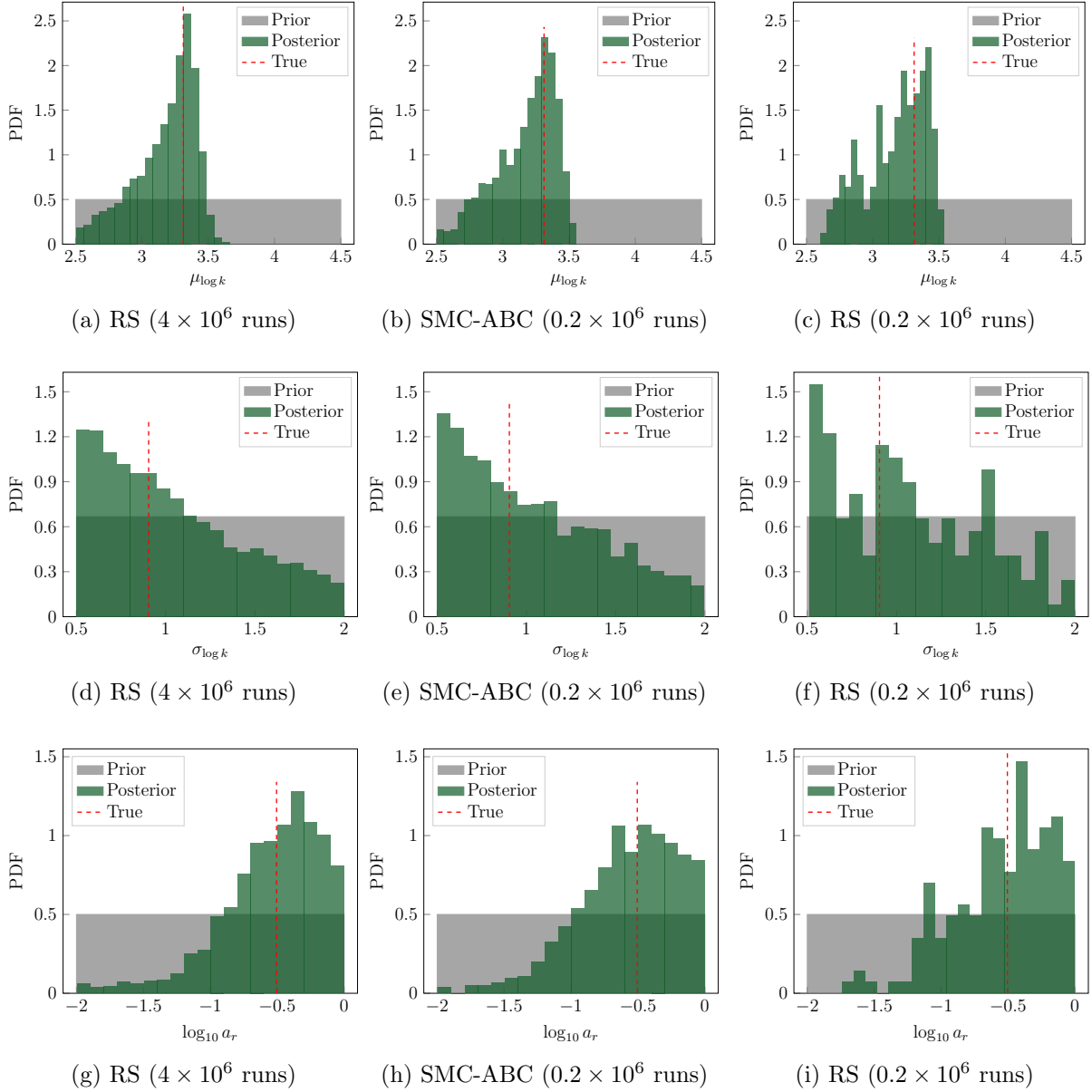

\centering
\begin{subfigure}[t]{.33\textwidth}
  \centering
  \resizebox{\linewidth}{!}{\input{Figures/marginal_posterior/RS4_perm_mean}}
  \caption{RS ($4 \times10^6$ runs)}
  \label{fig:subplot_a}
\end{subfigure}%
\hfill 
\begin{subfigure}[t]{.33\textwidth}
  \centering
  \resizebox{\linewidth}{!}{\input{Figures/marginal_posterior/smc_perm_mean}}
  \caption{SMC-ABC ($0.2 \times10^6$ runs)}
  \label{fig:subplot_b}
\end{subfigure}
\hfill 
\begin{subfigure}[t]{.33\textwidth}
  \centering
  \resizebox{\linewidth}{!}{\input{Figures/marginal_posterior/RS0.2_perm_mean}}
  \caption{RS ($0.2 \times10^6$ runs)}
  \label{fig:subplot_c}
\end{subfigure}

\vspace{5 mm}

\begin{subfigure}[t]{.33\textwidth}
  \centering
  \resizebox{\linewidth}{!}{\input{Figures/marginal_posterior/RS4_perm_std}}
  \caption{RS ($4 \times10^6$ runs)}
  \label{fig:subplot_a}
\end{subfigure}%
\hfill 
\begin{subfigure}[t]{.33\textwidth}
  \centering
  \resizebox{\linewidth}{!}{\input{Figures/marginal_posterior/smc_perm_std}}
  \caption{SMC-ABC ($0.2 \times10^6$ runs)}
  \label{fig:subplot_b}
\end{subfigure}
\hfill 
\begin{subfigure}[t]{.33\textwidth}
  \centering
  \resizebox{\linewidth}{!}{\input{Figures/marginal_posterior/RS0.2_perm_std}}
  \caption{RS ($0.2 \times10^6$ runs)}
  \label{fig:subplot_c}
\end{subfigure}

\vspace{5mm}

\begin{subfigure}[t]{.33\textwidth}
  \centering
  \resizebox{\linewidth}{!}{\input{Figures/marginal_posterior/RS4_kzkx}}
  \caption{RS ($4 \times10^6$ runs)}
  \label{fig:subplot_a}
\end{subfigure}%
\hfill 
\begin{subfigure}[t]{.33\textwidth}
  \centering
  \resizebox{\linewidth}{!}{\input{Figures/marginal_posterior/smc_kzkx}}
  \caption{SMC-ABC ($0.2 \times10^6$ runs)}
  \label{fig:subplot_b}
\end{subfigure}
\hfill 
\begin{subfigure}[t]{.33\textwidth}
  \centering
  \resizebox{\linewidth}{!}{\input{Figures/marginal_posterior/RS0.2_kzkx}}
  \caption{RS ($0.2 \times10^6$ runs)}
  \label{fig:subplot_c}
\end{subfigure}
\caption{Marginal posterior distributions of hyperparameters for true model~1. The left column shows the posterior distributions using RS with $4 \times 10^6$ forward runs, while the middle and right columns show the posterior distributions using SMC-ABC and RS, each with $0.2 \times 10^6$ runs. Gray regions represent the prior distribution, green histograms are the posterior distribution, and dashed red lines denote the true values.
}
\label{fig:marginal_posterior}
\end{figure}

We next present results for hyperparameter estimates. In Fig.~\ref{fig:marginal_posterior}, the marginal posterior distributions for the three hyperparameters using SMC-ABC and RS are compared. The left column displays the reference posterior results (RS with $4\times10^6$ forward runs), while the middle and right columns show the posterior distributions using the SMC-ABC and RS methods, each with $0.2\times10^6$ runs. In these figures, the gray regions represent the prior distribution, green histograms are the posterior distribution, and vertical dashed lines denote the true hyperparameter values. Although the SMC-ABC distributions are slightly ‘noisier’ than the reference results, they capture the key features quite well. The RS results with $0.2\times10^6$ runs, by contrast, show significant differences relative to the reference results. These differences will be shown to be even more extreme for true model~2.

\begin{figure}[htpb]
\centering
\begin{subfigure}[t]{.33\textwidth}
  \centering
  \resizebox{\linewidth}{!}{\includegraphics{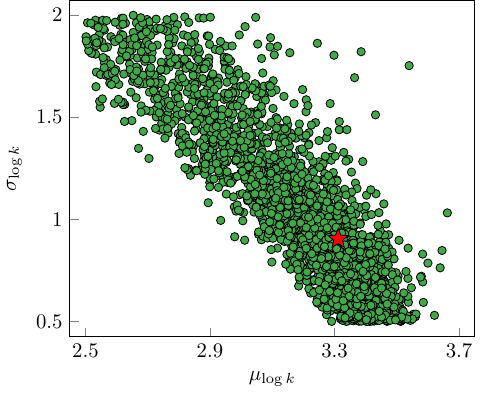}}
  \caption{RS ($4 \times10^6$ runs)}
  \label{fig:subplot_a}
\end{subfigure}%
\hfill 
\begin{subfigure}[t]{.33\textwidth}
  \centering
  \resizebox{\linewidth}{!}{\includegraphics{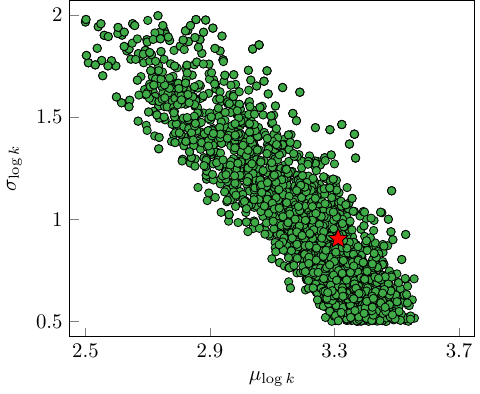}}
  \caption{SMC-ABC ($0.2 \times10^6$ runs)}
  \label{fig:subplot_b}
\end{subfigure}
\hfill 
\begin{subfigure}[t]{.33\textwidth}
  \centering
  \resizebox{\linewidth}{!}{\includegraphics{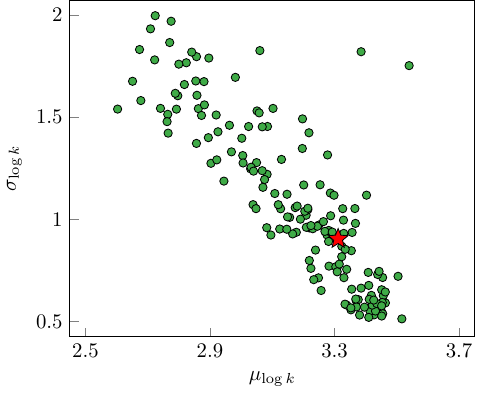}}
  \caption{RS ($0.2 \times10^6$ runs)}
  \label{fig:subplot_c}
\end{subfigure}

\vspace{5 mm}

\begin{subfigure}[t]{.33\textwidth}
  \centering
  \resizebox{\linewidth}{!}{\includegraphics{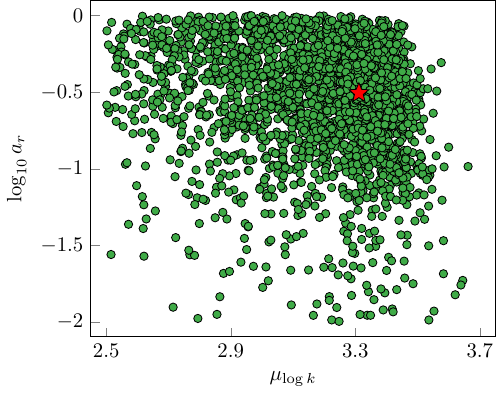}}
  \caption{RS ($4 \times10^6$ runs)}
  \label{fig:subplot_d}
\end{subfigure}%
\hfill 
\begin{subfigure}[t]{.33\textwidth}
  \centering
  \resizebox{\linewidth}{!}{\includegraphics{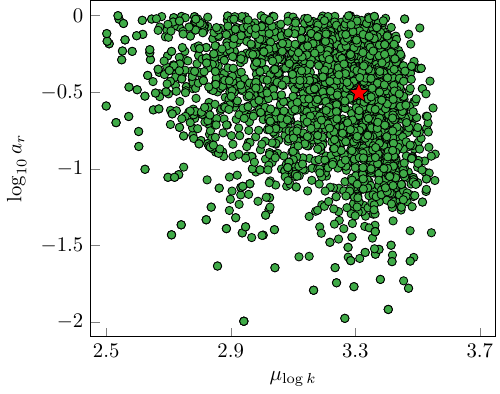}}
  \caption{SMC-ABC ($0.2 \times10^6$ runs)}
  \label{fig:subplot_e}
\end{subfigure}
\hfill 
\begin{subfigure}[t]{.33\textwidth}
  \centering
  \resizebox{\linewidth}{!}{\includegraphics{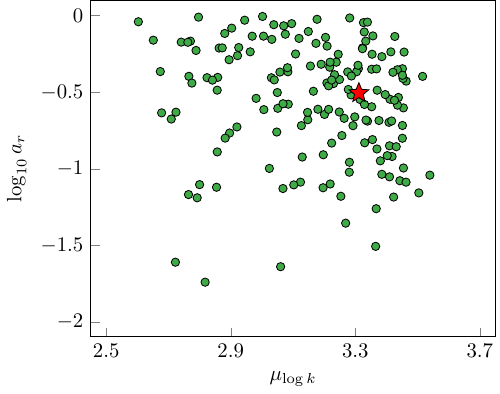}}
  \caption{RS ($0.2 \times10^6$ runs)}
  \label{fig:subplot_f}
\end{subfigure}

\vspace{5 mm}

\begin{subfigure}[t]{.33\textwidth}
  \centering
  \resizebox{\linewidth}{!}{\includegraphics{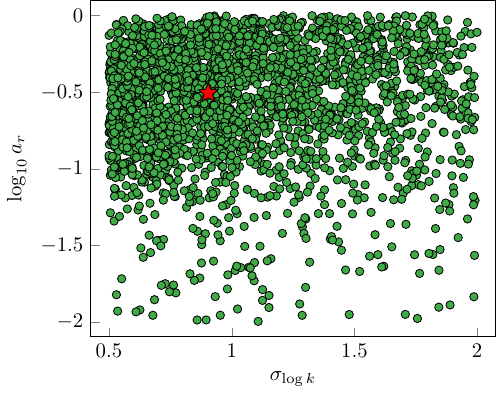}}
  \caption{RS ($4 \times10^6$ runs)}
  \label{fig:subplot_h}
\end{subfigure}%
\hfill 
\begin{subfigure}[t]{.33\textwidth}
  \centering
  \resizebox{\linewidth}{!}{\includegraphics{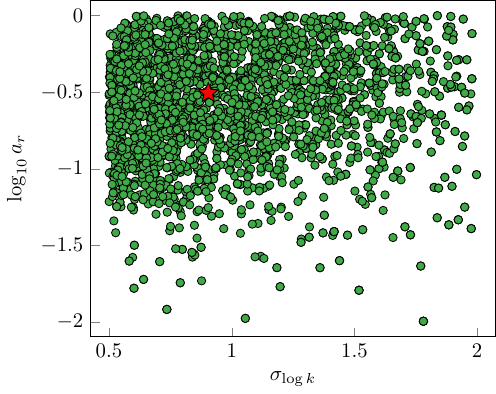}}
  \caption{SMC-ABC ($0.2 \times10^6$ runs)}
  \label{fig:subplot_i}
\end{subfigure}
\hfill 
\begin{subfigure}[t]{.33\textwidth}
  \centering
  \resizebox{\linewidth}{!}{\includegraphics{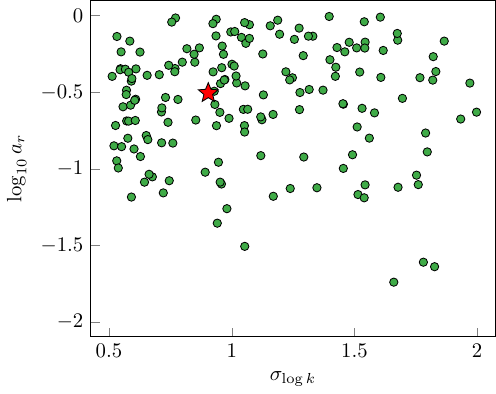}}
  \caption{RS ($0.2 \times10^6$ runs)}
  \label{fig:subplot_j}
\end{subfigure}
\caption{Pairwise marginal posterior samples of hyperparameters. Top row shows joint posterior samples of mean and standard deviation, middle row displays joint posterior samples of mean and permeability anisotropy ratio, and bottom row shows joint posterior samples of standard deviation and permeability anisotropy ratio. Green points represent posterior samples, and red stars denote true values.}
\label{fig:pairwise_marginal}
\end{figure}

The pairwise marginal posterior samples obtained by the two algorithms are shown in Fig.~\ref{fig:pairwise_marginal}. In each scatter plot, the green circles represent the joint values of two hyperparameters, and the red star indicates the true values. We see again that the SMC-ABC method with $0.2\times10^6$ runs provides posterior hyperparameter samples very close to those from RS with $4\times10^6$ runs. The use of RS with $0.2\times10^6$ runs yields sparsely distributed joint samples (due to the very low acceptance rate of RS). These scatter plots can also provide valuable insights for posterior hyperparameter values. It is clear from Fig.~\ref{fig:pairwise_marginal}(a) and~(b), for example, that realizations corresponding to small values of both $\mu_{\log k}$ and $\sigma_{\log k}$ (or large values of both hyperparameters) would not be expected to provide simulation results in agreement with observations.

The second stage of our hierarchical data assimilation approach involves the use of ESMDA to estimate the permeability field given a set of posterior hyperparameters and observed data. As noted in Section~\ref{sec:esmda}, 10 full ESMDA runs are performed, which corresponds to a total of 20,000 forward evaluations, in this second stage. The total computational cost for the hierarchical data assimilation procedure is thus $0.22\times10^6$ runs. By contrast, the converged rejection sampling requires $4\times10^6$ function evaluations (though reasonable results could be achieved with about half this number of runs). In either case, the hierarchical procedure provides an order of magnitude or more speedup for true model~1. As we will see, greater speedup is achieved for true model~2.

\begin{figure}[!h]
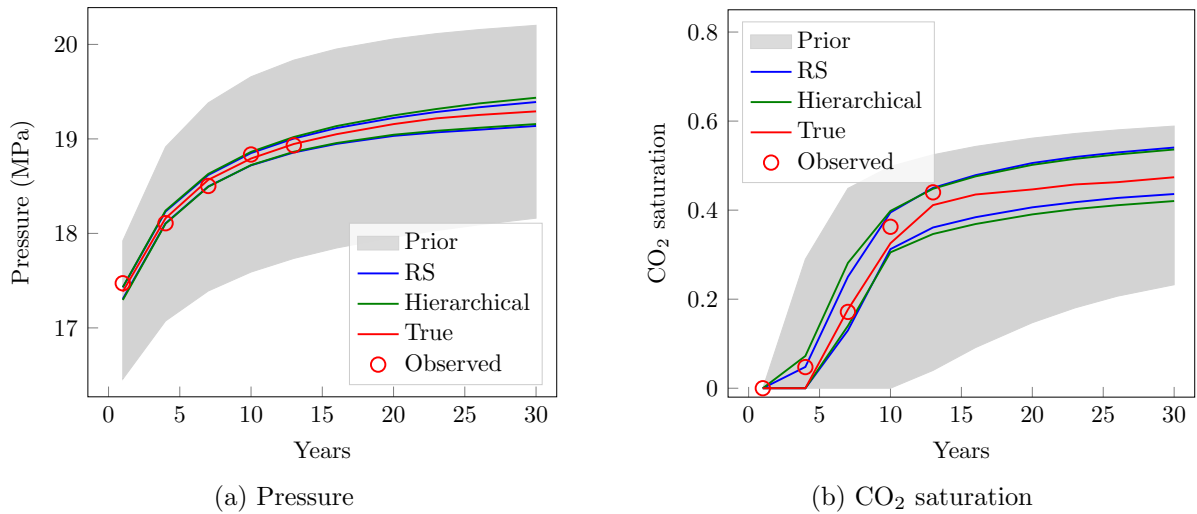

\centering
\begin{subfigure}{.47\textwidth}
  \centering
  \resizebox{\linewidth}{!}{\input{Figures/posterior_prediction/posterior_pres_case1_comparison}}
  \caption{Pressure}
  \label{fig:subplot_b}
\end{subfigure}
\hfill
\begin{subfigure}{.47\textwidth}
  \centering
  \resizebox{\linewidth}{!}{\input{Figures/posterior_prediction/posterior_sat_case1_comparison}}
  \caption{\( \text{CO}_{2} \) saturation}
  \label{fig:subplot_a}
\end{subfigure}%
\caption{History matching results for pressure (left) and \( \text{CO}_{2} \) saturation (right) at the monitoring location for true model 1. Gray regions represent the prior $\mathrm{P}_{10}$--$\mathrm{P}_{90}$ range, red lines denote true data, and red circles are observed data. Solid blue and green curves denote the posterior $\mathrm{P}_{10}$ (lower) and $\mathrm{P}_{90}$ (upper) predictions obtained using the RS and hierarchical procedure, respectively.}
\label{fig:posterior_prediction}
\end{figure}

History matching results for pressure and $\mathrm{CO}_{2}$ saturation at the monitoring location are shown in Fig.~\ref{fig:posterior_prediction}. The gray regions represent the prior $\mathrm{P}_{10}$--$\mathrm{P}_{90}$ range, the red lines are the true data generated using the surrogate model, and the red circles are observed data. The solid blue and green curves represent the posterior $\mathrm{P}_{10}$--$\mathrm{P}_{90}$ predictions obtained from the RS procedure (with $4\times10^6$ forward runs) and from the hierarchical approach, respectively. The hierarchical results are from the posterior models constructed from the 10 separate ESMDA runs. There is close agreement between the two sets of results, with substantial uncertainty reduction achieved for both pressure and saturation at the monitoring location. We see slight discrepancies between the two sets of results (e.g., $\mathrm{P}_{90}$ saturation at 5~years and $\mathrm{P}_{10}$ saturation from 10--30~years), though the differences are small.

We now comment on the timings using RS and SMC-ABC-ESMDA. The time required for a surrogate model forward run (to provide pressure and saturation) is about 0.06~seconds on a single Nvidia Tesla V100 GPU. The time needed to generate a geomodel using GSLIB (AMD EPYC 7502 32-core processor) is 0.24~seconds, so this is actually the more time-consuming step. The serial time to generate and evaluate $10^6$ models is, therefore, about 300,000~seconds, or about 83~hours. The times required for RS and SMC-ABC-ESMDA computations differ slightly, but the main component is model generation and forward model evaluation. Thus, RS with $4 \times 10^6$ models requires about 333~hours of serial computation, while SMC-ABC-ESMDA with $0.22 \times 10^6$ models requires about 18.3~hours of serial computation. Note that the SMC-ABC-ESMDA workflow with ECLIPSE runs (which take 3~minutes each) would entail about 11,000~hours of serial computation. The training time using 2000 ECLIPSE runs corresponds to a serial time of 100~hours for simulation (AMD EPYC 7502 32-core processor), and a total of 26~hours for network training (on a V100 GPU). The timings given here for training and for the various workflows can be reduced significantly using parallelization or job arrays. Nonetheless, it is clear that the use of the surrogate model is essential for our data assimilation workflows.

\subsection{Data assimilation results for true model~2}

\begin{figure}[!h]
\centering
\begin{subfigure}[t]{.33\textwidth}
  \centering
  \resizebox{\linewidth}{!}{\includegraphics{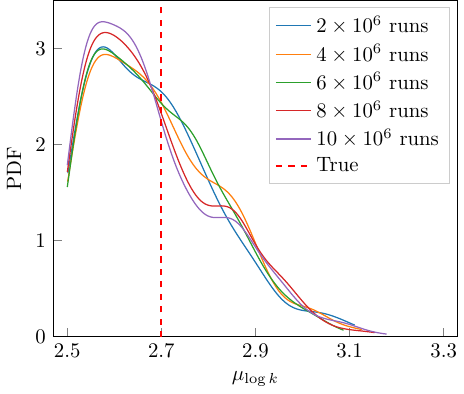}}
  \caption{$\mu_{\log k}$}
  \label{fig:subplot_a}
\end{subfigure}%
\hfill
\begin{subfigure}[t]{.33\textwidth}
  \centering
  \resizebox{\linewidth}{!}{\includegraphics{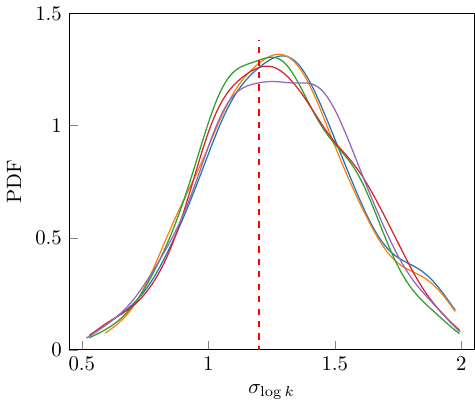}}
  \caption{$\sigma_{\log k}$}
  \label{fig:subplot_b}
\end{subfigure}
\hfill 
\begin{subfigure}[t]{.33\textwidth}
  \centering
  \resizebox{\linewidth}{!}{\includegraphics{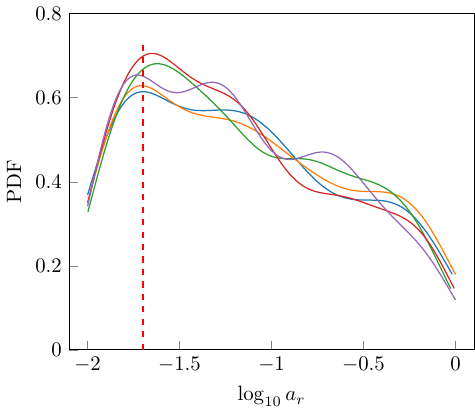}}
  \caption{$\log_{10}a_r$}
  \label{fig:subplot_c}
\end{subfigure}
\caption{Marginal posterior distributions of hyperparameters obtained from RS for true model~2. Solid curves represent the posterior distributions obtained using different numbers
of forward runs. Dashed vertical lines denote the true values of the hyperparameters. Legend in (a) applies to all subplots.}
\label{fig:case2_reference}
\end{figure}

In the interest of brevity, we present a more limited set of results for true model~2. The posterior distributions of the hyperparameters for this case, obtained using the RS method, are shown in Fig.~\ref{fig:case2_reference}. This example is more challenging than true model~1, presumably because the true response falls at the edge of the prior. As a result, we continue to see small changes in the distributions even as we proceed from $8\times10^6$ to $10\times10^6$ forward runs. In fact, after $10\times10^6$ runs, only 773 samples have been accepted by RS (this acceptance rate is almost an order of magnitude lower than for true model~1). Although slight shifts may still occur with larger numbers of function evaluations, we take the RS results with $10\times10^6$ runs to be the reference results.

The population size in SMC-ABC for true model~2 is set to 1000, which is roughly consistent with the number of accepted samples (773) from RS using $10\times10^6$ runs. The behavior of the JS divergence with the number of forward runs for true model~2 (though not shown) is similar to that for true model~1. We use the SMC-ABC results derived from $0.1\times10^6$ forward runs to compare with the reference RS results, as this case provides a near-zero JS divergence relative to RS with $10\times10^6$ runs.

\begin{figure}[H]
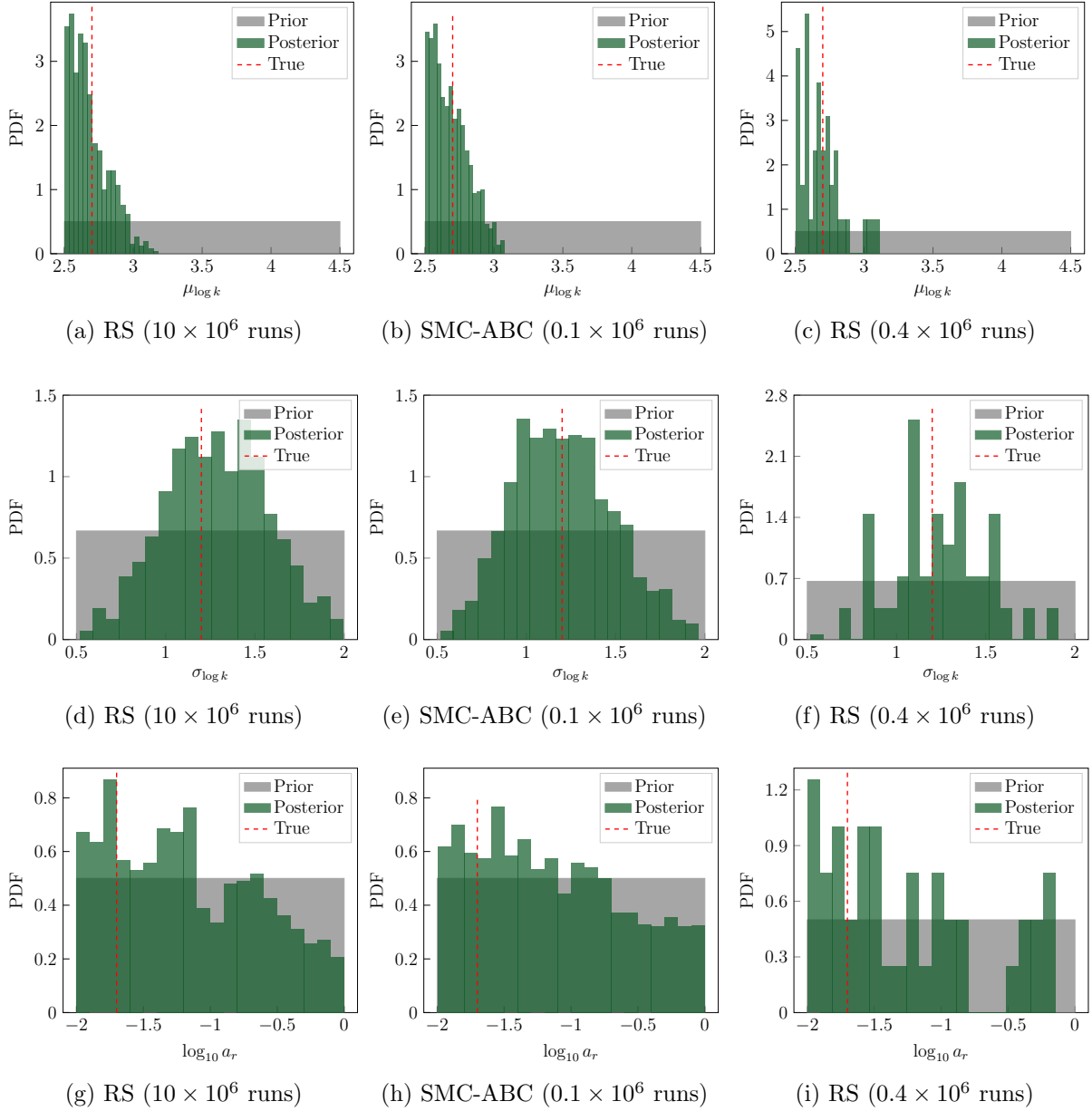

\centering
\begin{subfigure}[t]{.33\textwidth}
  \centering
  \resizebox{\linewidth}{!}{\input{Figures/case2_marginal/RS10_perm_mean_case2}}
  \caption{RS ($10 \times10^6$ runs)}
  \label{fig:subplot_a}
\end{subfigure}%
\hfill 
\begin{subfigure}[t]{.33\textwidth}
  \centering
  \resizebox{\linewidth}{!}{\input{Figures/case2_marginal/smc_perm_mean_case2}}
  \caption{SMC-ABC ($0.1 \times10^6$ runs)}
  \label{fig:subplot_b}
\end{subfigure}
\hfill 
\begin{subfigure}[t]{.33\textwidth}
  \centering
  \resizebox{\linewidth}{!}{\input{Figures/case2_marginal/RS0.4_perm_mean_case2}}
  \caption{RS ($0.4 \times10^6$ runs)}
  \label{fig:subplot_c}
\end{subfigure}

\vspace{5 mm}

\begin{subfigure}[t]{.33\textwidth}
  \centering
  \resizebox{\linewidth}{!}{\input{Figures/case2_marginal/RS10_perm_std_case2}}
  \caption{RS ($10 \times10^6$ runs)}
  \label{fig:subplot_a}
\end{subfigure}%
\hfill 
\begin{subfigure}[t]{.33\textwidth}
  \centering
  \resizebox{\linewidth}{!}{\input{Figures/case2_marginal/smc_perm_std_case2}}
  \caption{SMC-ABC ($0.1 \times10^6$ runs)}
  \label{fig:subplot_b}
\end{subfigure}
\hfill 
\begin{subfigure}[t]{.33\textwidth}
  \centering
  \resizebox{\linewidth}{!}{\input{Figures/case2_marginal/RS0.4_perm_std_case2}}
  \caption{RS ($0.4 \times10^6$ runs)}
  \label{fig:subplot_c}
\end{subfigure}

\vspace{5 mm}

\begin{subfigure}[t]{.33\textwidth}
  \centering
  \resizebox{\linewidth}{!}{\input{Figures/case2_marginal/RS10_kzkx_case2}}
  \caption{RS ($10 \times10^6$ runs)}
  \label{fig:subplot_a}
\end{subfigure}%
\hfill 
\begin{subfigure}[t]{.33\textwidth}
  \centering
  \resizebox{\linewidth}{!}{\input{Figures/case2_marginal/smc_kzkx_case2}}
  \caption{SMC-ABC ($0.1 \times 10^6$ runs)}
  \label{fig:subplot_b}
\end{subfigure}
\hfill 
\begin{subfigure}[t]{.33\textwidth}
  \centering
  \resizebox{\linewidth}{!}{\input{Figures/case2_marginal/RS0.4_kzkx_case2}}
  \caption{RS ($0.4 \times10^6$ runs)}
  \label{fig:subplot_c}
\end{subfigure}
\caption{Marginal posterior distributions of hyperparameters for true model~2. The left column shows the posterior distributions using RS with $10 \times 10^6$ forward runs, the middle column displays the posterior distributions using SMC-ABC with $0.1 \times 10^6$ runs, and the right column presents the posterior distributions using RS with $0.4 \times 10^6$ runs. Gray regions represent the prior distribution, green histograms are the posterior distribution, and dashed red lines denote the true values.}
\label{fig:case2_marginal}
\end{figure}

The marginal posterior distributions of hyperparameters using RS and SMC-ABC are presented in Fig.~\ref{fig:case2_marginal}. A large uncertainty reduction is observed for $\mu_{\log k}$, and some reduction is achieved for $\sigma_{\log k}$. It is evident that the SMC-ABC method, with $0.1\times10^6$ runs, can produce similar marginal posterior distributions for the hyperparameters as RS with $10\times10^6$ runs. RS with $0.4\times10^6$ runs, by contrast, results in poorly resolved approximations, as only 43~samples are accepted in this case.

History matching results for pressure and saturation at the monitoring location are shown in Fig.~\ref{fig:case2_prediction}. We again observe significant uncertainty reduction and reasonable agreement in the $\mathrm{P}_{10}$ and $\mathrm{P}_{90}$ posterior predictions. Note that the $\mathrm{P}_{10}$--$\mathrm{P}_{90}$ posterior range is slightly larger with the hierarchical procedure than with RS. This was also observed in Fig.~\ref{fig:posterior_prediction}. Further runs would be required to determine if results shift, with either approach, with more function evaluations. In any event, our results demonstrate that the hierarchical data assimilation approach is able to generate posterior distributions and predictions that are in close correspondence with the much more expensive RS method.

\begin{figure}[!h]
\centering
\begin{subfigure}[t]{.49\textwidth}
  \centering
  \resizebox{\linewidth}{!}{\input{Figures/case2_prediction/posterior_pres_comparison}}
  \caption{Pressure}
  \label{fig:subplot_b}
\end{subfigure}
\hfill
\begin{subfigure}[t]{.49\textwidth}
  \centering
  \resizebox{\linewidth}{!}{\input{Figures/case2_prediction/posterior_sat_comparison}}
  \caption{\( \text{CO}_{2} \) saturation}
  \label{fig:subplot_a}
\end{subfigure}%
\caption{History matching results for pressure (left) and \( \text{CO}_{2} \) saturation (right) at the monitoring location for true model~2. Gray regions represent the prior $\mathrm{P}_{10}$--$\mathrm{P}_{90}$ range, red lines denote true data, and red circles are observed data. Solid blue and green curves denote the posterior $\mathrm{P}_{10}$ (lower) and $\mathrm{P}_{90}$ (upper) predictions obtained using RS and the hierarchical procedure, respectively.}
\label{fig:case2_prediction}
\end{figure}

Computational requirements for both true models, in terms of forward runs, are summarized in Table~\ref{table:computational_cost}. The hierarchical data assimilation method achieves a speedup of $83\times$ for true model~2, which is significantly larger than that for true model~1. This increased speedup is due to both an increase in the number of runs required by RS and a decrease in the number of runs for SMC-ABC (due to the smaller population size). RS is, in general, inefficient because it uses ‘brute-force’ prior sampling in the hyperparameter space to match the observed data in the data space. In the case of true model~2, because the observed pressure data fall beyond the $\mathrm{P}_{90}$ of the prior distribution, RS is particularly inefficient (i.e., even less efficient than for true model~1). The SMC-ABC approach, by contrast, focuses its computational effort on hyperparameter values that are likely to produce close matches to the observed data, which leads to substantial efficiency gains.

\begin{table}[H]
\small
\setstretch{1.5}
\centering
\caption{Computational cost in terms of forward runs (each run entails a geostatistical
simulation and a surrogate function evaluation) using RS and SMC-ABC-ESMDA, for both true models.}
\label{table:computational_cost}
\begin{tabular}{p{3.1cm} p{3.1cm} p{4.7cm} p{3.1cm}}
\hline
\noalign{\hrule height 0.8pt}
 & RS & SMC-ABC-ESMDA & Speedup \\ 
\hline
True model~1 & $4\times10^6$ & $0.22\times10^6$ & $18\times$\\ 
True model~2 & $10\times10^6$ & $0.12\times10^6$ & $83\times$\\ 
\hline
\noalign{\hrule height 0.8pt}
\end{tabular}
\end{table}
\section{Comparison to Standalone ESMDA}
\label{sec:Traditional Data Assimilation}

Ensemble Kalman-based methods are widely used in practice, so it is appropriate to compare our procedure with this type of approach. Here we consider an ensemble smoother with multiple data assimilation (ESMDA). We first describe the modifications to the standard ESMDA procedure necessary to enable hyperparameter estimation. Posterior results are then presented and compared to those obtained using the reference RS and our hierarchical data assimilation procedure.

\subsection{Modified ESMDA procedure}

ESMDA is commonly applied with fixed hyperparameters (as in the second step of our SMC-ABC-ESMDA methodology). To enable the estimation of posterior hyperparameters using only an ESMDA procedure, several changes are introduced in Algorithm~\ref{esmda}. Specifically, the initial ensemble of permeability realizations is constructed by first sampling hyperparameters from the hyperprior distributions. A prior realization based on the sampled hyperparameters is then generated using GSLIB. In addition, the state vector in ESMDA is augmented to include the permeability anisotropy ratio as an additional parameter. Finally, the posterior distributions of the mean and standard deviation of log-permeability are computed directly from the posterior ESMDA permeability fields.

In general, the performance of ESMDA can potentially be improved by increasing the ensemble size and the number of data assimilation steps. This is because a large ensemble size can reduce sampling errors and additional assimilation steps can improve convergence. In this work, we conduct multiple ESMDA runs, corresponding to different computational budgets, to evaluate ESMDA performance with increasing function evaluations. The ensemble size $N_e$, number of assimilation steps $N_a$, and inflation coefficients $\alpha_j$ used for each ESMDA run are given in Table~\ref{table:esmda_setup}. The specific $\alpha_j$ for $N_a = 4$ and 10 are those suggested in~\cite{emerickInvestigation2013}. The $\alpha_j$ for $N_a = 20$ were determined heuristically, while satisfying the constraint $\sum_{j=1}^{N_a}\alpha_j^{-1} = 1$. The computational costs in terms of surrogate model evaluations for each case are 2000, 4000, 25,000, 50,000, 100,000, 150,000, and 200,000. We use true model~1, described in Section~\ref{sec:results}, in this assessment.

\begin{table}[htpb]
\small
\setstretch{1.5}
\centering
\caption{Standalone ESMDA specifications corresponding to different computational budgets}
\label{table:esmda_setup}
\begin{tabular}{p{2.6cm} p{2.6cm} p{9.5cm}}
\hline
\noalign{\hrule height 0.8pt}
Ensemble size & Iterations & Inflation coefficients \\ 
\hline
500 & 4 & [9.333, 7.0, 4.0, 2.0] \\ 
1000 & 4 & [9.333, 7.0, 4.0, 2.0] \\ 
2500 & 10 & [57.017, 35.0, 25.0, 20.0, 18.0, 15.0, 12.0, 8.0, 5.0, 3.0] \\ 
5000 & 10 & [57.017, 35.0, 25.0, 20.0, 18.0, 15.0, 12.0, 8.0, 5.0, 3.0] \\ 
10,000 & 10 & [57.017, 35.0, 25.0, 20.0, 18.0, 15.0, 12.0, 8.0, 5.0, 3.0] \\ 
7500 & 20 & [129.635, 105.0, 95.0, 85.0, 75.0, 65.0, 60.0, 55.0, 50.0, 45.0, 40.0, 35.0, 30.0, 25.0, 20.0, 15.0, 12.0, 9.0, 6.0, 4.0] \\ 
10,000 & 20 & [129.635, 105.0, 95.0, 85.0, 75.0, 65.0, 60.0, 55.0, 50.0, 45.0, 40.0, 35.0, 30.0, 25.0, 20.0, 15.0, 12.0, 9.0, 6.0, 4.0] \\ 
\hline
\noalign{\hrule height 0.8pt}
\end{tabular}
\end{table}

\subsection{Data assimilation results}

We first present results in terms of Jensen-Shannon (JS) divergence, described in Section~\ref{sec:true model 1}. This metric quantifies the accuracy of the hyperparameter estimation relative to the reference RS results. JS divergence results for standalone ESMDA (black curves) are shown in Fig.~\ref{fig:esmda_js}. The SMC-ABC results (red curves, these are the same as in Fig.~\ref{fig:efficiency_comparison}) are also displayed. We see that, for two of the hyperparameters (Fig.~\ref{fig:esmda_js}(a) and (c)), ESMDA provides lower JS divergence for (relatively) small numbers of function evaluations. However, the ESMDA results do not display proper convergence behavior, meaning JS divergence does not continue to decrease with increasing numbers of function evaluations, for any of the three hyperparameters. This demonstrates that, although effective in many cases, standalone ESMDA may display an inherent bias in hyperparameter estimation.

This bias results in part from the ensemble-based approximation of the covariance matrix. In traditional ensemble-based data assimilation, all prior realizations are constructed from the same covariance matrix, and the ESMDA objective function is defined under this assumption (as shown in Eq.~\eqref{objective}). In our problem, however, the covariance matrix depends on the (uncertain) hyperparameters. This dependency causes the ensemble-based approximation of covariance to lose accuracy. In addition, in this work we use uniform hyperprior distributions, and this is also incompatible with underlying (Gaussian) assumptions in Kalman-based data assimilation methods.

\begin{figure}[!ht]
\centering
\begin{subfigure}[t]{.33\textwidth}
  \centering
  \resizebox{\linewidth}{!}{\input{Figures/esmda_comparison/esmda_js_mean}}
  \caption{$\mu_{\log k}$}
  \label{fig:subplot_a}
\end{subfigure}%
\hfill 
\begin{subfigure}[t]{.33\textwidth}
  \centering
  \resizebox{\linewidth}{!}{\input{Figures/esmda_comparison/esmda_js_std}}
  \caption{$\sigma_{\log k}$}
  \label{fig:subplot_b}
\end{subfigure}
\hfill 
\begin{subfigure}[t]{.33\textwidth}
  \centering
  \resizebox{\linewidth}{!}{\input{Figures/esmda_comparison/esmda_js_kzkx}}
  \caption{$\log_{10}a_r$}
  \label{fig:subplot_c}
\end{subfigure}
\caption{Convergence comparison of hyperparameter estimation using SMC-ABC and the modified ESMDA.}
\label{fig:esmda_js}
\end{figure}

For an appropriate comparison of modified ESMDA relative to our hierarchical data assimilation framework (SMC-ABC-ESMDA), we consider results corresponding to similar computational budgets. As discussed in Section~\ref{sec:true model 1}, SMC-ABC with around 200,000 forward runs provides posterior results in close agreement with those from the reference RS. We thus use the standalone ESMDA run with an ensemble size of 10,000 and 20 assimilation steps (200,000 function evaluations) in the following comparisons. The serial computation time in this case is comparable to that using SMC-ABC-ESMDA.

The marginal posterior distributions of hyperparameters using the modified ESMDA are shown in Fig.~\ref{fig:esmda_marginal}. The ESMDA estimates show substantial differences compared to the reference (RS and SMC-ABC) results in Fig.~\ref{fig:marginal_posterior}. In particular, with ESMDA, no uncertainty reduction is achieved for $\sigma_{\log k}$. In addition, Gaussian-like results, including positive values, are observed for $\log_{10}a_r$. These behaviors are in contrast to the reference results in Fig.~\ref{fig:marginal_posterior}. We thus conclude that even with a comparable number of function evaluations, this ESMDA method is unable to provide `correct' posterior distributions of hyperparameters. We note that ESMDA is still able to give practically useful results, and that it may be possible to develop ESMDA variants with better performance in terms of these posterior distributions.

\begin{figure}[!ht]
\centering
\begin{subfigure}[t]{.33\textwidth}
  \centering
  \resizebox{\linewidth}{!}{\input{Figures/esmda_comparison/esmda_mu}}
  \caption{$\mu_{\log k}$}
  \label{fig:subplot_a}
\end{subfigure}%
\hfill 
\begin{subfigure}[t]{.33\textwidth}
  \centering
  \resizebox{\linewidth}{!}{\input{Figures/esmda_comparison/esmda_sigma}}
  \caption{$\sigma_{\log k}$}
  \label{fig:subplot_b}
\end{subfigure}
\hfill 
\begin{subfigure}[t]{.33\textwidth}
  \centering
  \resizebox{\linewidth}{!}{\input{Figures/esmda_comparison/esmda_kzkx}}
  \caption{$\log_{10}a_r$}
  \label{fig:subplot_c}
\end{subfigure}
\caption{Marginal posterior distributions of hyperparameters obtained using the modified ESMDA procedure for true model~1 with 200,000 function evaluations. Reference results from RS and SMC-ABC are shown in Fig.~\ref{fig:marginal_posterior}. Gray regions represent the prior distribution, green histograms are the posterior distribution, and dashed red lines denote the true values.}
\label{fig:esmda_marginal}
\end{figure}

History matching results obtained using the modified ESMDA procedure are shown in Fig.~\ref{fig:esmda_posterior_prediction}. The solid blue and dashed black curves represent the posterior $\mathrm{P}_{10}$--$\mathrm{P}_{90}$ predictions obtained from RS and the modified ESMDA approach, respectively. The results show that ESMDA produces a much larger $\mathrm{P}_{10}$--$\mathrm{P}_{90}$ range of pressure predictions than RS, though the differences in posterior saturation predictions are small. The lower $\mathrm{P}_{10}$ result for pressure with ESMDA may be related to the overestimation of $\log_{10}a_r$ in Fig.~\ref{fig:esmda_marginal}(c).

\begin{figure}[!h]
\centering
\begin{subfigure}{.48\textwidth}
  \centering
  \resizebox{\linewidth}{!}{\input{Figures/esmda_comparison/esmda_posterior_pres}}
  \caption{Pressure}
  \label{fig:subplot_b}
\end{subfigure}
\hfill
\begin{subfigure}{.48\textwidth}
  \centering
  \resizebox{\linewidth}{!}{\input{Figures/esmda_comparison/esmda_posterior_sat}}
  \caption{\( \text{CO}_{2} \) saturation}
  \label{fig:subplot_a}
\end{subfigure}%
\caption{History matching results for pressure (left) and \( \text{CO}_{2} \) saturation (right) at the monitoring location for true model 1. Gray regions represent the prior $\mathrm{P}_{10}$--$\mathrm{P}_{90}$ range, red lines denote true data, and red circles are observed data. Solid blue and dashed black curves denote the posterior $\mathrm{P}_{10}$ (lower) and $\mathrm{P}_{90}$ (upper) predictions obtained using RS and the modified ESMDA procedure (with 200,000 function evaluations), respectively.}
\label{fig:esmda_posterior_prediction}
\end{figure}

\begin{figure}[H]
\centering
\begin{subfigure}[t]{.31\textwidth} 
  \centering
  \resizebox{\linewidth}{!}{\includegraphics{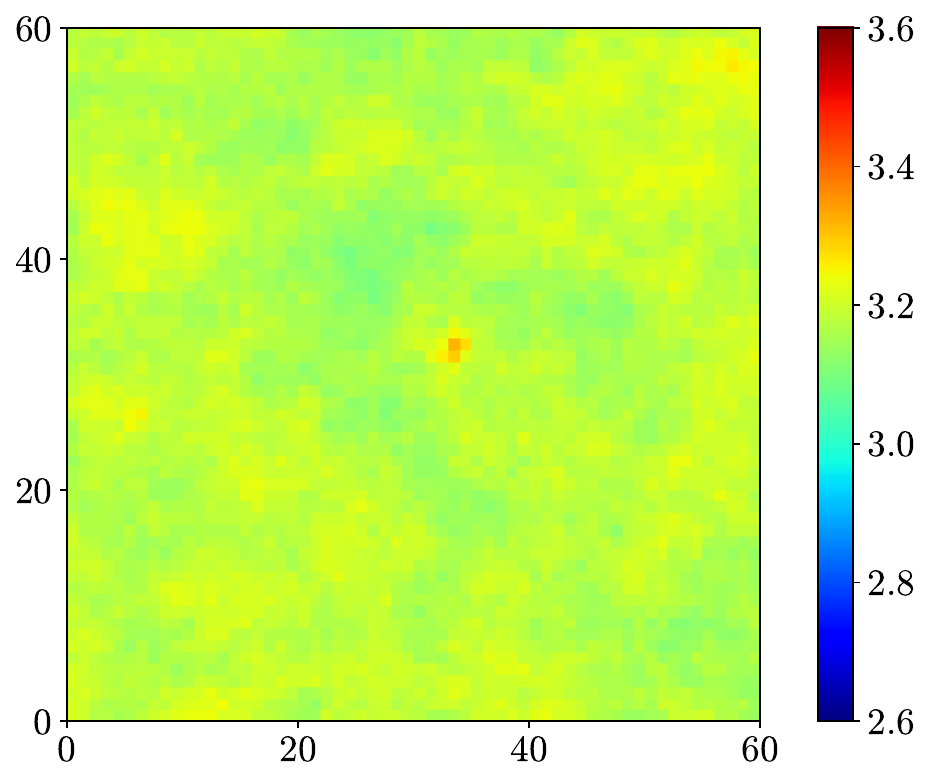}}
  \caption{Post.~mean (RS)}
  \label{fig:subplot_a}
\end{subfigure}%
\hfill
\begin{subfigure}[t]{.31\textwidth}
  \centering
  \resizebox{\linewidth}{!}{\includegraphics{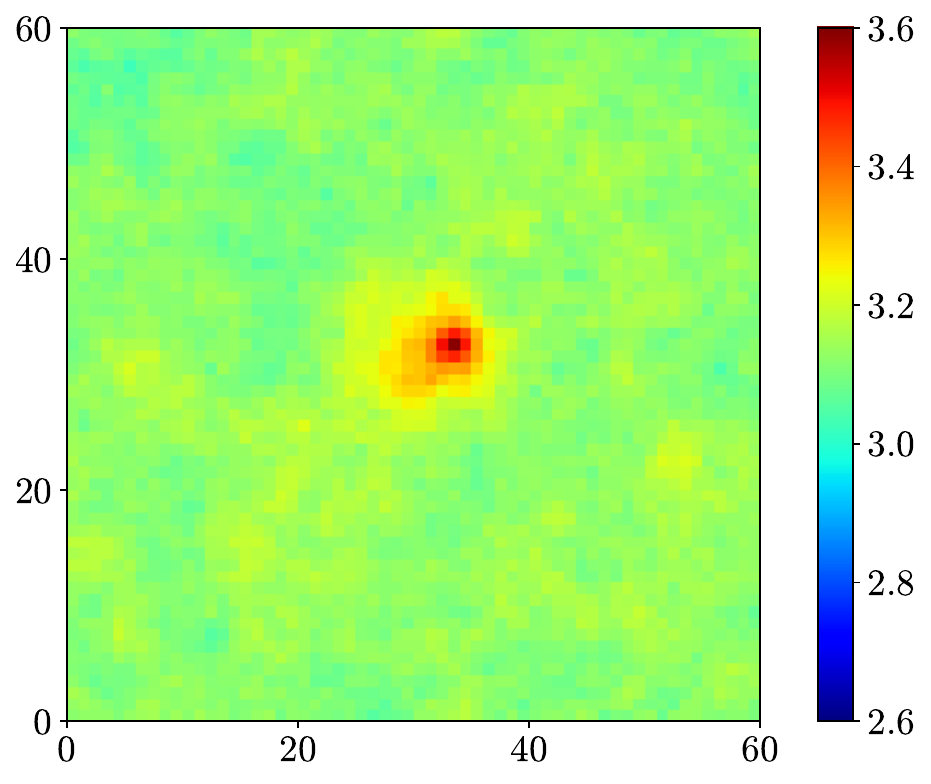}}
  \caption{Post.~mean (SMC-ABC-ESMDA)}
  \label{fig:subplot_b}
\end{subfigure}
\hfill
\begin{subfigure}[t]{.31\textwidth}
  \centering
  \resizebox{\linewidth}{!}{\includegraphics{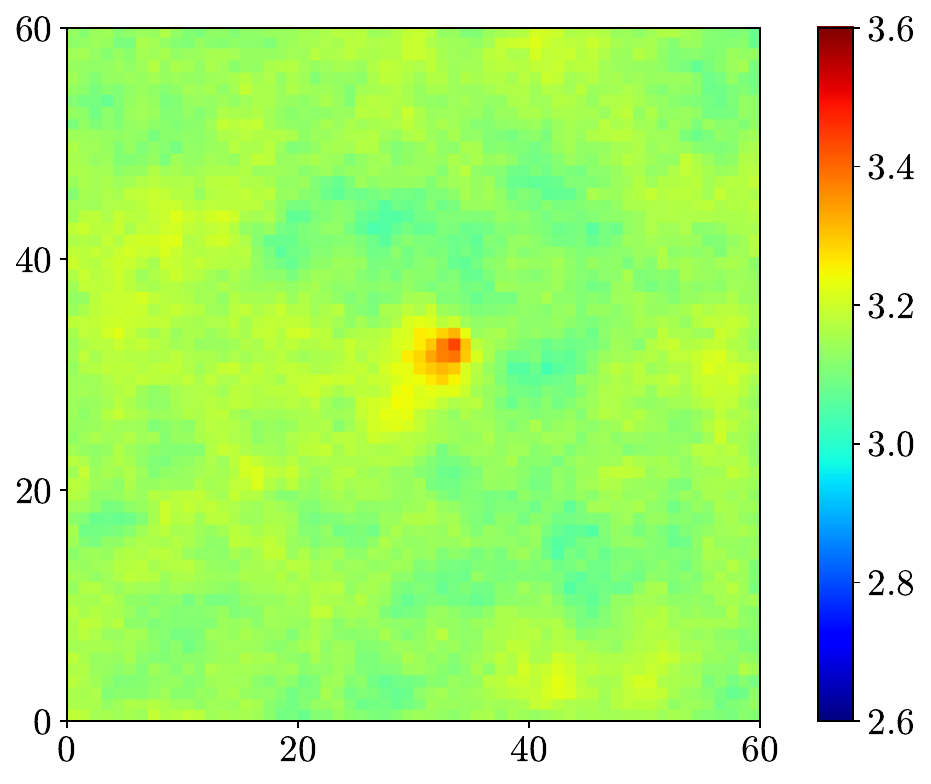}}
  \caption{Post.~mean (ESMDA)}
  \label{fig:subplot_c}
\end{subfigure}

\vspace{0.5 mm}

\begin{subfigure}[t]{.31\textwidth}
  \centering
  \resizebox{\linewidth}{!}{\includegraphics{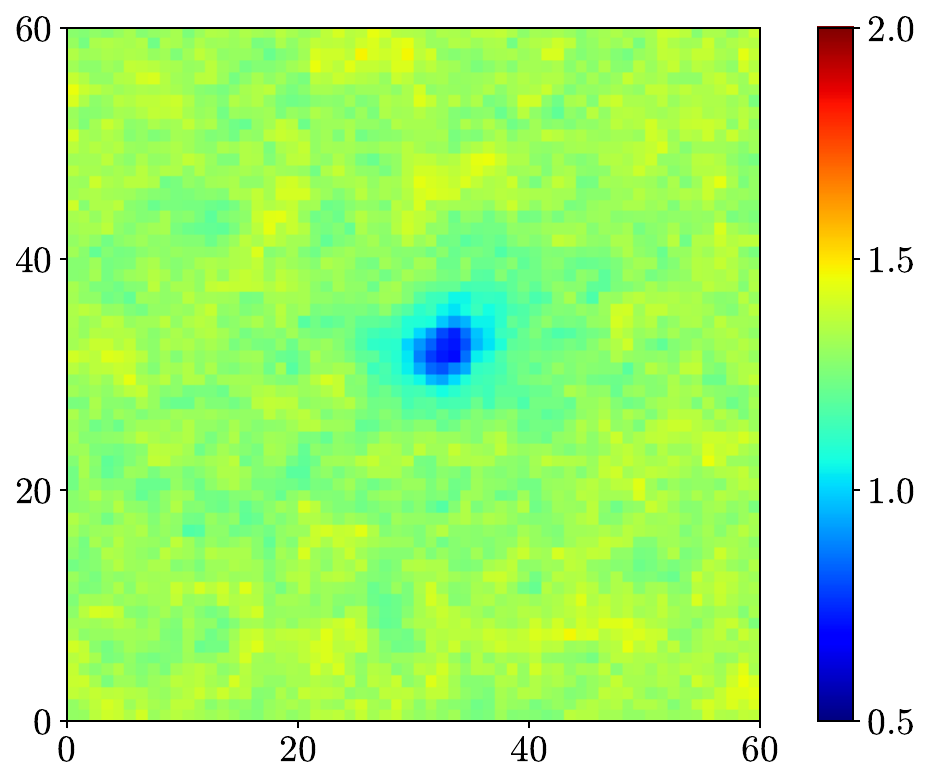}}
  \caption{Post.~variance (RS)}
  \label{fig:subplot_d}
\end{subfigure}%
\hfill
\begin{subfigure}[t]{.31\textwidth}
  \centering
  \resizebox{\linewidth}{!}{\includegraphics{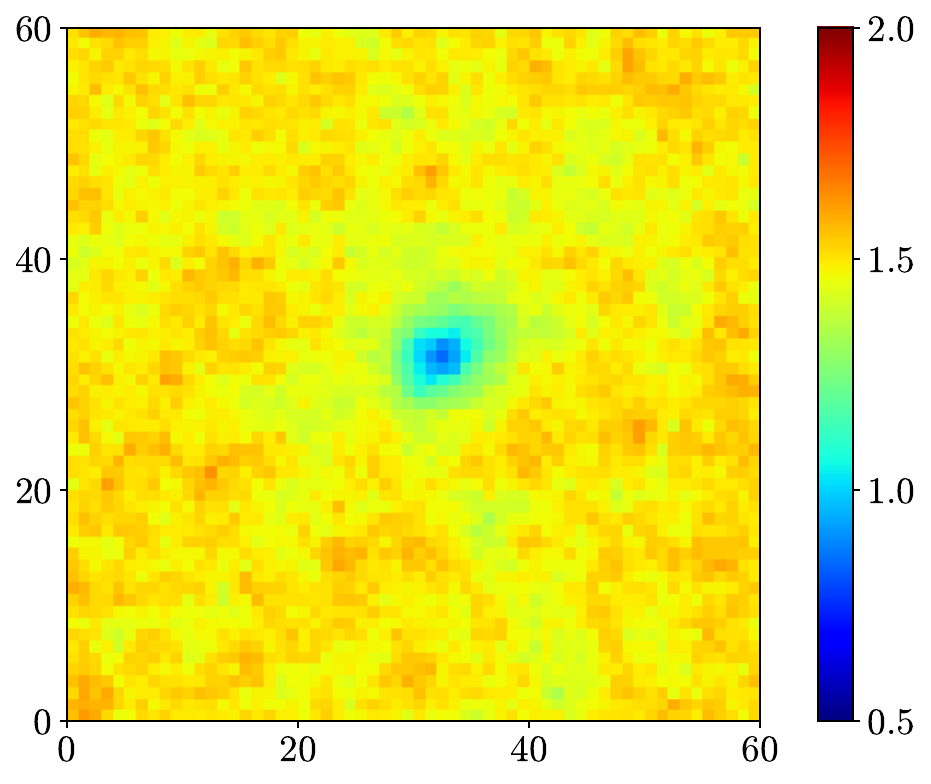}}
  \caption{Post.~variance (SMC-ABC-ESMDA)}
  \label{fig:subplot_e}
\end{subfigure}
\hfill
\begin{subfigure}[t]{.31\textwidth}
  \centering
  \resizebox{\linewidth}{!}{\includegraphics{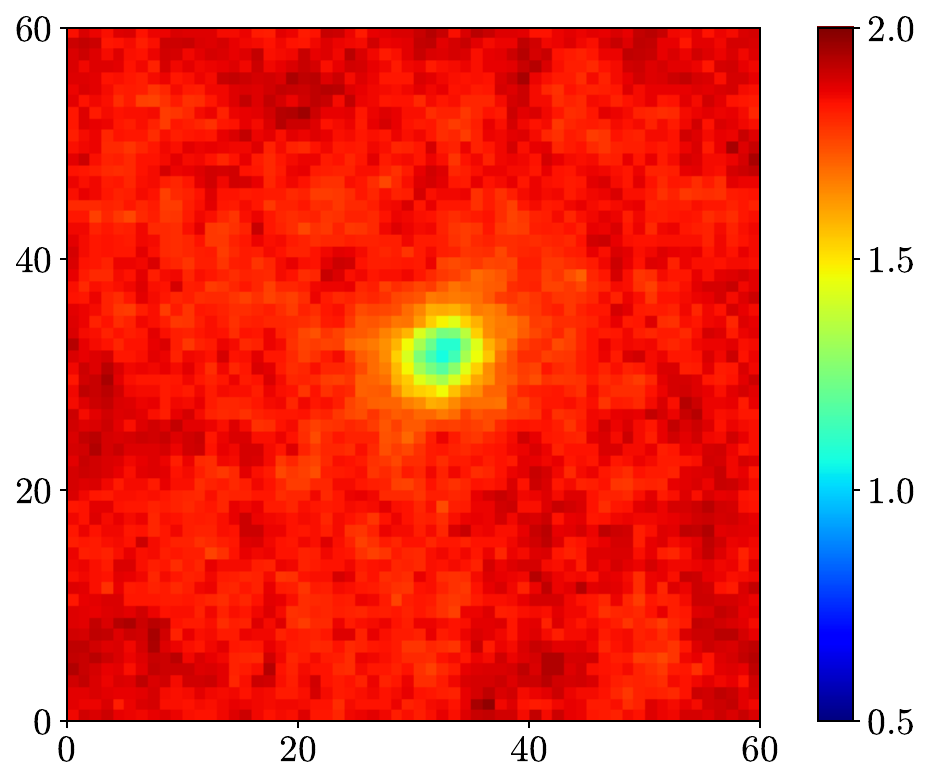}}
  \caption{Post.~variance (ESMDA)}
  \label{fig:subplot_f}
\end{subfigure}

\vspace{0.5 mm}

\begin{subfigure}[t]{.31\textwidth}
  \centering
  \resizebox{\linewidth}{!}{\includegraphics{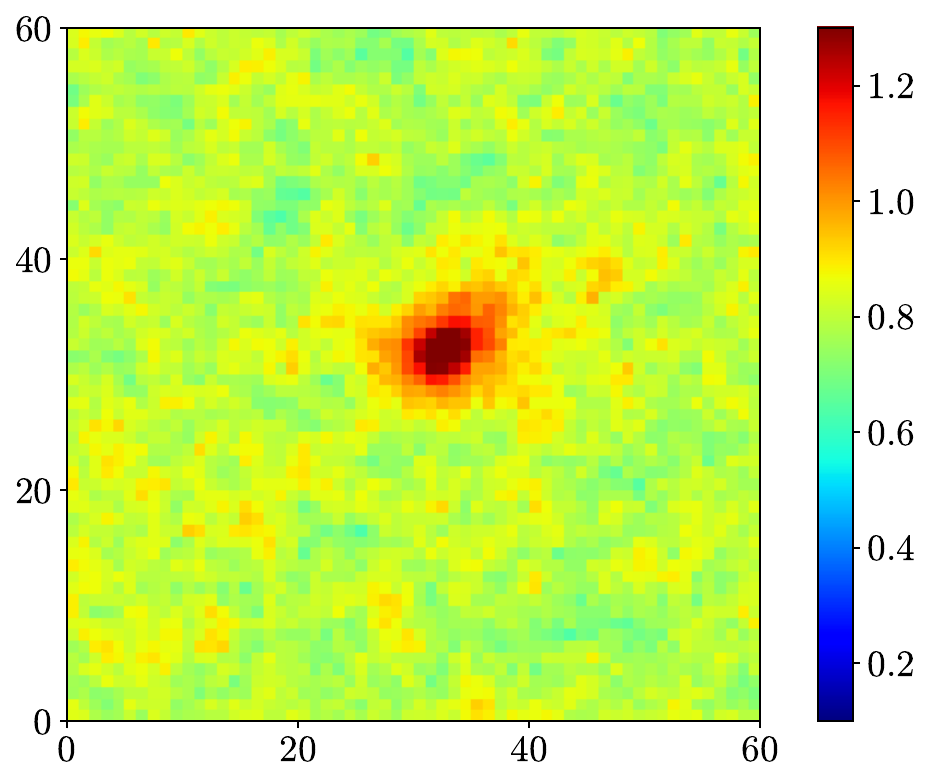}}
  \caption{Var.~reduction (RS)}
  \label{fig:subplot_g}
\end{subfigure}%
\hfill
\begin{subfigure}[t]{.31\textwidth}
  \centering
  \resizebox{\linewidth}{!}{\includegraphics{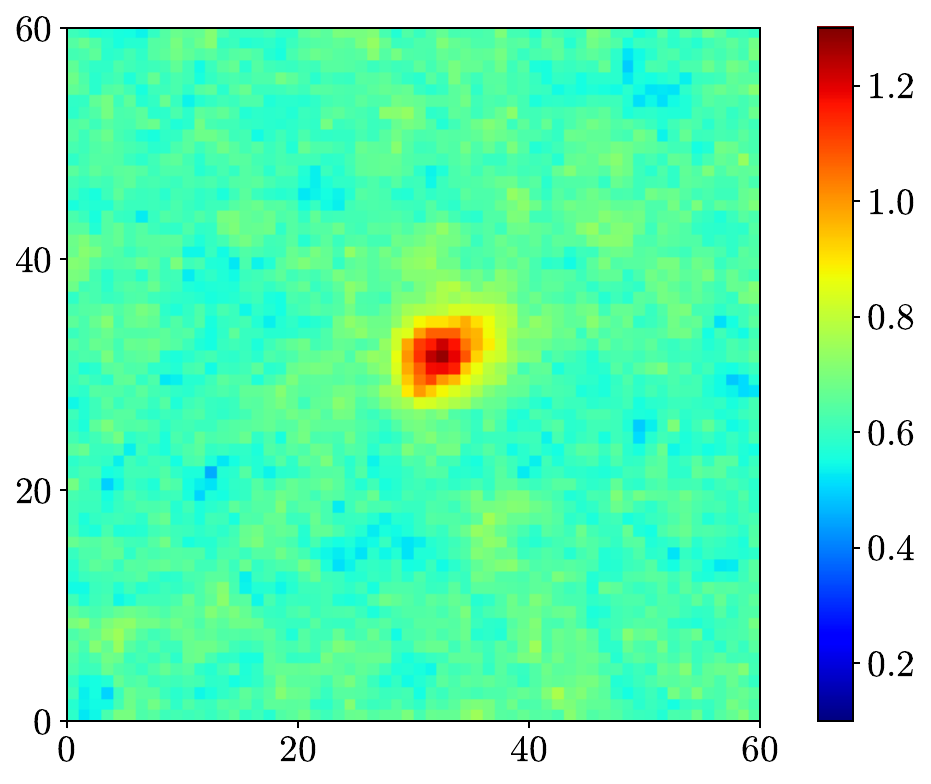}}
  \caption{Var.~reduction (SMC-ABC-ESMDA)}
  \label{fig:subplot_h}
\end{subfigure}
\hfill
\begin{subfigure}[t]{.31\textwidth}
  \centering
  \resizebox{\linewidth}{!}{\includegraphics{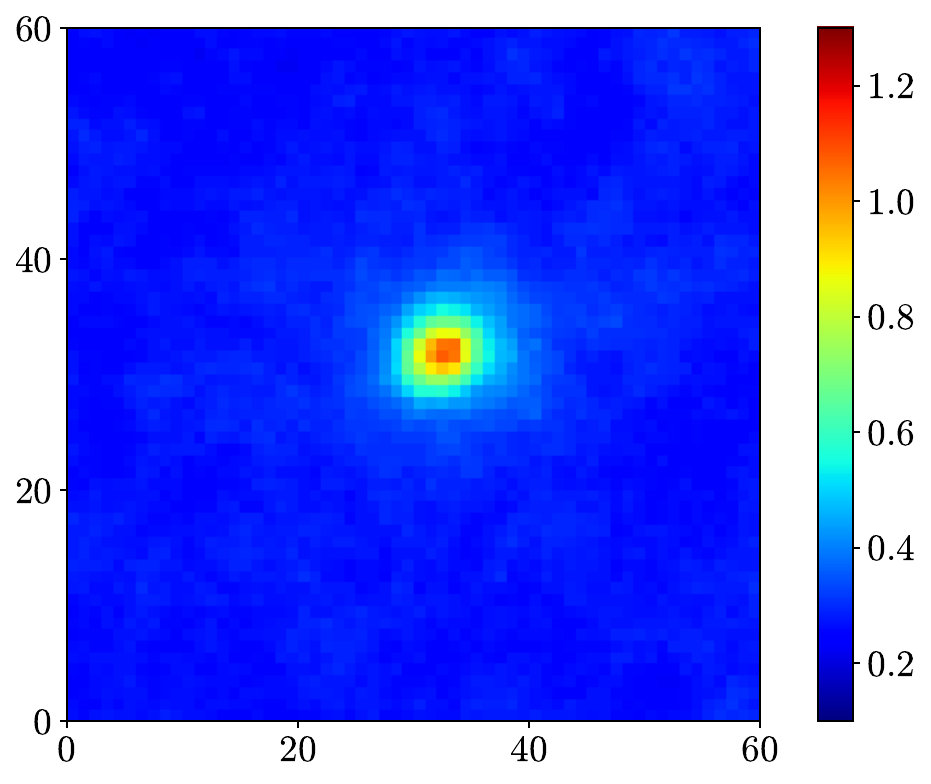}}
  \caption{Var.~reduction (ESMDA)}
  \label{fig:subplot_i}
\end{subfigure}

\vspace{0.5 mm}

\begin{subfigure}[t]{.31\textwidth}
  \centering
  \resizebox{\linewidth}{!}{\includegraphics{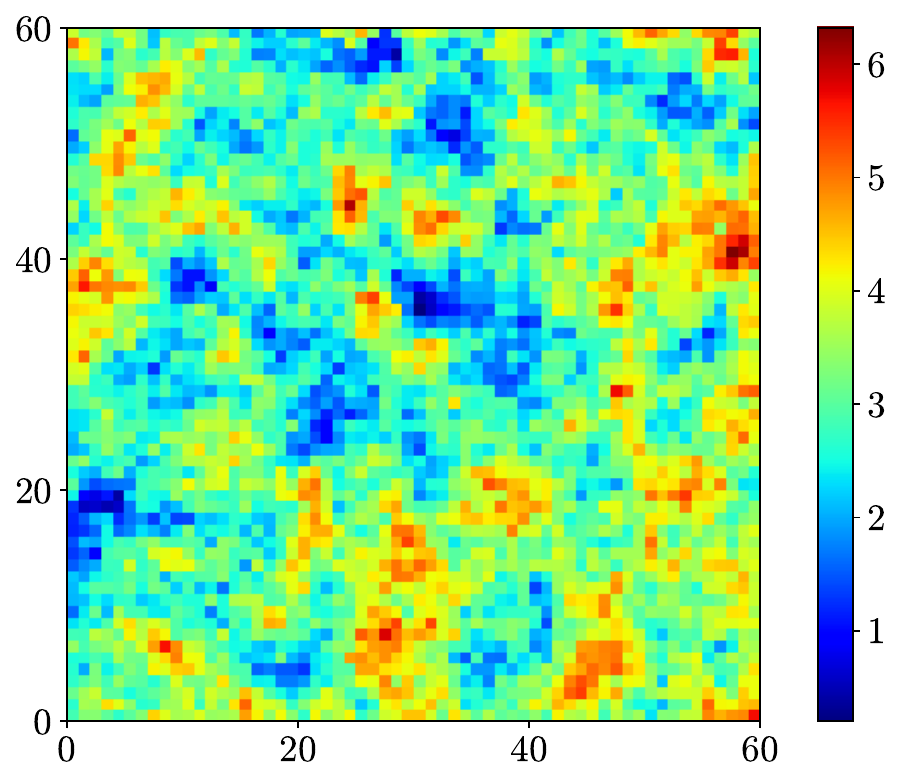}}
  \caption{True log-permeability}
  \label{fig:subplot_j}
\end{subfigure}
\caption{Posterior results for log-permeability in the top layer of the model obtained using RS, SMC-ABC-ESMDA, and the modified ESMDA procedure (with 200,000 function evaluations). Top  row shows posterior mean, second row displays posterior variance, and third row depicts variance reduction. True log-permeability in the top layer shown in (j).}
\label{fig:perm_comparison}
\end{figure}

Finally, we display results for posterior log-permeability mean and variance obtained using RS, SMC-ABC-ESMDA, and the modified ESMDA procedure. These results, for the top layer of the model, are shown in Fig.~\ref{fig:perm_comparison}. As described in Section~\ref{sec:hierarchical}, the posterior permeability fields from the hierarchical approach are generated with 10 (separate) standard ESMDA runs using 10 representative sets of posterior hyperparameters, while those from RS are obtained based on all joint posterior samples of hyperparameters and permeability fields. The posterior mean results (top row) are in general correspondence, in terms of scale, between all three methods. It should be noted that, given the high dimension of the permeability field (43,200 grid blocks) and the limited amount of observed data used for history matching, none of these results is expected to reproduce the true model (shown in Fig.~\ref{fig:perm_comparison}(j)).

The posterior variance and variance reduction results, in the second and third rows of Fig.~\ref{fig:perm_comparison}, show approximate correspondence between RS and SMC-ABC-ESMDA. More specifically, away from the well, variance for RS (Fig.~\ref{fig:perm_comparison}(d)) is about 1.3-1.5, while for SMC-ABC-ESMDA (Fig.~\ref{fig:perm_comparison}(e)) it is about 1.4-1.6. The variance with modified ESMDA is noticeably larger -- around 1.6-2.0 in the region away from the well. This is consistent with the posterior distribution in Fig.~\ref{fig:esmda_marginal}(b), where we see essentially no uncertainty reduction in $\sigma_{\log k}$.

As noted above, the SMC-ABC-ESMDA results in Fig.~\ref{fig:perm_comparison} are generated by combining the results for 10 separate ESMDA runs (with hyperparameters sampled from the converged SMC-ABC run). The posterior mean and variance results differ between these ESMDA runs because they each correspond to a different set of hyperparameters. In some cases, rather than generate a set of combined results (as in Fig.~\ref{fig:perm_comparison}), it may be instructive to consider posterior mean and variance for each set of ESMDA hyperparameters.

\section{Concluding Remarks}
\label{sec:conclusion}

In this study, we developed and applied a hierarchical data assimilation framework for geological carbon storage problems. The framework uses the SMC-ABC method to estimate geomodel hyperparameters, followed by ESMDA to provide posterior realizations of grid-block permeability. An existing 3D recurrent R-U-Net deep learning-based surrogate model was applied for function evaluations. This is an essential component of the workflow, as the overall methodology would be computationally prohibitive if high-fidelity flow simulations were used. The accuracy of the surrogate model for our case, which involved the injection of 1~Mt/year of CO$_2$ (for 30~years) through a single injector in the center of a 3D model, was demonstrated through comparisons to high-fidelity simulation results.

Detailed data assimilation results, including quantitative comparisons to a reference RS procedure, were presented for two synthetic true models. The marginal posterior distributions of the hyperparameters, pairwise marginal posterior samples, and history matching results for pressure and saturation time-series at the monitoring location were compared. Close agreement in the posterior results from the two approaches was demonstrated for both true models. This is significant, since the converged RS computations required $4\times10^6$ (true model~1) and $10\times10^6$ (true model~2) forward runs. Our hierarchical procedure, by contrast, needed $0.22\times10^6$ and $0.12\times10^6$ runs for the two cases, respectively, corresponding to speedups of $18\times$ and $83\times$ compared to RS. Thus our findings clearly demonstrate that the SMC-ABC-ESMDA procedure is able to generate accurate posterior results, with reasonable computational efficiency, for the cases considered.

We also compared the performance of a modified (standalone) ESMDA procedure to that of SMC-ABC-ESMDA. The results show that standalone ESMDA indeed provides uncertainty reduction, but the posterior distributions of the hyperparameters, and the posterior predictions of pressure at the monitoring location, show clear discrepancies with reference RS results. In addition, the JS divergence of the ESMDA results with increasing number of function evaluations does not display proper convergence behavior. It may be possible to devise a standalone ESMDA procedure that achieves better performance, though this can only be confirmed by comparison against reference results.

There are many promising directions for future research in this area. From an algorithmic perspective, it will be of interest to improve the sampling efficiency using an active learning strategy. The integration of Bayesian optimization with likelihood-free inference has the potential to reduce the number of forward simulations, and this should be explored. For hyperparameter estimation, normalizing flows, which seek an invertible mapping between a simple base distribution and the target posterior distribution, could be considered. In terms of applications, the hierarchical framework should be extended as required to treat realistic problems. Of particular interest are cases involving sequential decision making. This is important for geological carbon storage because, in practical settings, new injection wells will be drilled and new monitoring wells will be required as the operation proceeds. SMC-based algorithms are effective for sequential decisions, and the methodology developed in this paper should be extended for this case.

\section*{CRediT authorship contribution statement}
\textbf{Wenchao Teng}: Conceptualization, Methodology, Software, Formal analysis, Visualization, Writing -- Original Draft. \textbf{Louis J.~Durlofsky}: Conceptualization, Formal analysis, Writing -- Review and Editing, Supervision.

\section*{Declaration of competing interest}
The authors declare that they have no known competing financial interests or personal relationships that could have appeared to influence the work reported in this paper.

\section*{Data availability}
Please contact Wenchao Teng (wenchaot@stanford.edu) for the data or code used in this study.

\section*{Acknowledgments}
We are grateful to the Stanford Graduate Fellowship in Science and Engineering and the Stanford Smart Fields Consortium for financial support. We thank Meng Tang and Yifu Han for providing the recurrent R-U-Net surrogate model code, and Tapan Mukerji, Peter Kitanidis, Catherine Gorle, Dylan Crain, Xiaowen He, Amy Zou and Haoyu Tang for useful discussions. We acknowledge the SDSS Center for Computation for HPC resources.

\bibliographystyle{elsarticle-harv}
\bibliography{references}
\end{document}